\pdfoutput=1

\documentclass[11pt]{article}

\usepackage[]{acl}

\usepackage{times}
\usepackage{latexsym}

\usepackage[T1]{fontenc}

\usepackage[utf8]{inputenc}

\usepackage{microtype}

\usepackage{inconsolata}

\usepackage{graphicx}

\usepackage{amsmath}
\usepackage{algorithm}
\usepackage{algpseudocode}
\usepackage{url}
\usepackage{multirow}
\usepackage{enumitem}
\usepackage{array}
\usepackage[symbol]{footmisc}
\usepackage{threeparttable}
\usepackage{dblfloatfix} 
\usepackage{caption}
\usepackage[labelformat=simple]{subcaption}

\usepackage{adjustbox}
\usepackage{multirow}
\usepackage{hyperref}
\usepackage{float}
\usepackage{longtable}

\usepackage{booktabs}
\usepackage{colortbl}

\definecolor{LightGreen}{rgb}{0.88, 1.0, 0.88}
\definecolor{LightBlue}{rgb}{0.88, 0.92, 1.0}

\usepackage{amsmath}
\usepackage{amssymb}
\usepackage{tabularx}
\usepackage{makecell}
\usepackage{mathtools}

\newcolumntype{P}[1]{>{\centering\arraybackslash}p{#1}}
\usepackage{tabularx}
\usepackage[graphicx]{realboxes}
\usepackage[most]{tcolorbox}
\usepackage{soul} 
\usepackage{xcolor}
\usepackage{array}
\usepackage{tikz}
\usetikzlibrary{shapes.geometric, arrows.meta, positioning}

\usepackage{blindtext}
\newcommand\blfootnote[1]{%
\begingroup
\renewcommand\thefootnote{}\footnote{#1}%
\addtocounter{footnote}{-1}%
\endgroup
}

\title{MedCOD: Enhancing English-to-Spanish Medical Translation of Large Language Models Using Enriched Chain-of-Dictionary Framework}

\author{
Md Shahidul Salim$^{1,2}$, 
Lian Fu$^3$, 
Arav Adikesh Ramakrishnan$^3$,
Zonghai Yao\thanks{Co-corresponding authors}$^{1,3}$, 
Hong Yu\footnotemark[1]$^{1,2,3}$\\
$^{1}$Center for Healthcare Organization and Implementation Research, VA Bedford Health Care  \\
$^{2}$Miner School of Computer and Information Sciences, University of Massachusetts Lowell \\
$^{3}$ Manning College of Information and Computer Sciences, University of Massachusetts Amherst\\
\texttt\ \url{MdShahidul\_Salim@student.uml.edu}
  \\
}

\begin{document}
\maketitle
\begin{abstract}
We present \textbf{MedCOD (Medical Chain-of-Dictionary)}, a hybrid framework designed to improve English-to-Spanish medical translation by integrating domain-specific structured knowledge into large language models (LLMs).
MedCOD integrates domain-specific knowledge from both the Unified Medical Language System (UMLS) and the LLM-as-Knowledge-Base (LLM-KB) paradigm to enhance structured prompting and fine-tuning.
We constructed a parallel corpus of 2,999 English–Spanish MedlinePlus articles and a 100-sentence test set annotated with structured medical contexts. Four open-source LLMs (Phi-4, Qwen2.5-14B, Qwen2.5-7B, and LLaMA-3.1-8B) were evaluated using structured prompts that incorporated multilingual variants, medical synonyms, and UMLS-derived definitions, combined with LoRA-based fine-tuning.
Experimental results demonstrate that MedCOD significantly improves translation quality across all models. 
For example, Phi-4 with MedCOD and fine-tuning achieved BLEU 44.23, chrF++ 28.91, and COMET 0.863, surpassing strong baseline models like GPT-4o and GPT-4o-mini.
Ablation studies confirm that both MedCOD prompting and model adaptation independently contribute to performance gains, with their combination yielding the highest improvements. 
These findings highlight the potential of structured knowledge integration to enhance LLMs for medical translation tasks.

\end{abstract}

\section{Introduction}

\blfootnote{$\ddagger$ To appear in Findings of the Association for Computational Linguistics: EMNLP 2025}

\begin{figure*}[!ht]
    \centering
    \includegraphics[width=\linewidth]{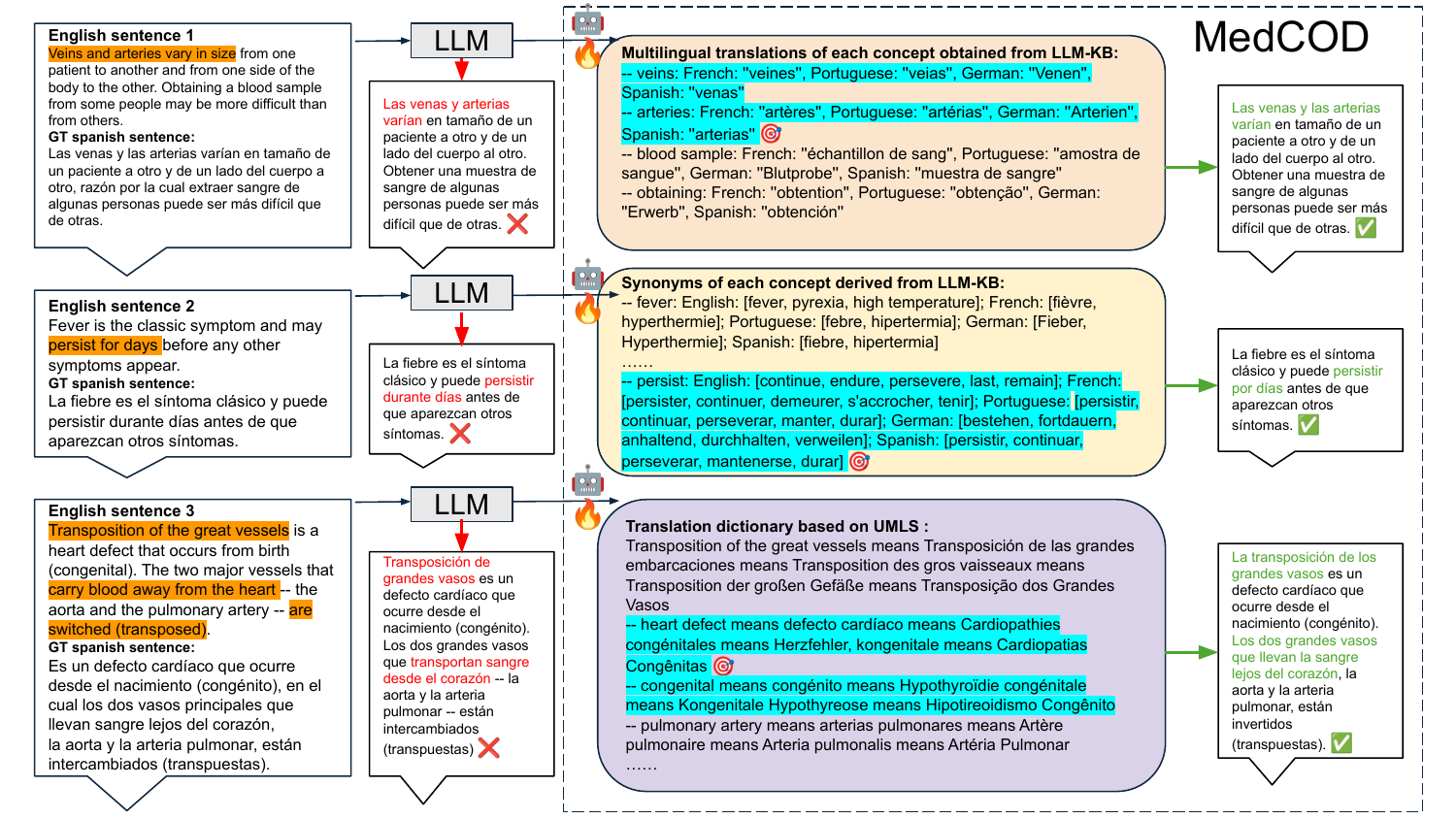}
    \caption{Overview of the MedCOD framework for improving English-to-Spanish medical translation.
    The figure illustrates three representative English medical sentences (left), their initial LLM-generated Spanish translations (middle), and how MedCOD enhances them (right) by utilizing structured domain knowledge. Specifically, MedCOD integrates:
    (a) multilingual term mappings from a large language model knowledge base (LLM-KB);
    (b) synonym expansion from LLM-KB to match concept variations; and
    (c) concept-aligned UMLS dictionary translations with disambiguation support.
    For example, Sentence 3 demonstrates MedCOD's ability to correct terminology by disambiguating "transposition of the great vessels" and “carry blood away from the heart,” improving grammaticality, semantics, and clinical accuracy, e.g., using "la transposición de los grandes vasos" and "llevan la sangre lejos del corazón." These enhancements ensure accurate, fluent, and medically faithful Spanish outputs.}
    \label{fig:medcod}
\end{figure*}

Electronic Health Records (EHRs) have become an integral part of modern healthcare, serving as a critical medium for enhancing patient engagement and facilitating better communication between healthcare providers and patients~\cite{delbanco2012inviting, gabay201721st,walker2019opennotes}.
Recognizing the value of EHRs, the Centers for Medicare \& Medicaid Services (CMS) Incentive Programs have promoted the meaningful use of EHRs, empowering patients to access and manage their health information electronically. 
However, the full benefits of EHR accessibility are not uniformly realized, particularly among patients with limited English proficiency~\cite{liu2015translating,root2016characteristics, kayastha2018open}.

As of 2021–2022, Hispanics account for approximately 19\% of the U.S. population (around 63 million people)~\cite{zong2022mosaic, Wikipedia2025}. Individuals with Limited English Proficiency (LEP) represent about 8\% of the total U.S. population, and Hispanics make up 62\% of the LEP group~\cite{haldar2023overview}. This translates to roughly 31–32\% of Hispanics having limited English skills.
Language barriers have a well-documented impact on health care access and comprehension. Nearly half of LEP adults report encountering challenges in the past three years, such as difficulty filling out medical forms (34\%), communicating with health professionals (33\%), understanding physician instructions (30\%), or following prescription guidelines (27\%)~\cite{haldar2023overview}.
Research further shows that LEP is associated with reduced medication adherence and poorer health outcomes, particularly among Hispanic patients with chronic conditions like asthma~\cite{mcquaid2018cultural}. Conversely, when patients are matched with language-concordant physicians, studies report significant improvements in satisfaction and chronic disease management outcomes~\cite{WikipediaLEP2025}.

To address these challenges, machine translation (MT) technologies have been explored for medical communication. Traditional statistical MT (SMT) and neural MT (NMT) systems~\cite{brown1990,riina2024evaluation} demonstrate promise but often struggle with domain-specific terminology, abbreviations, and context preservation. General-purpose MT systems such as Google Translate and Bing Translator, while widely accessible, frequently fail to ensure clinical accuracy in longer or complex sentences~\cite{liu2015translating, kapoor2022, turner2019}. Although LLMs have recently achieved remarkable advances in general-domain multilingual translation, their potential for translating biomedical texts, particularly EHRs, remains underexplored~\cite{riina2024evaluation}.

In this study, we propose a novel translation framework named \textbf{MedCOD}, which builds upon Chain-of-Dictionary Prompting (COD)~\cite{codpaper}. MedCOD integrates structured medical knowledge from the Unified Medical Language System (UMLS)~\cite{lindberg1993unified} and an LLM-as-Knowledge-Base (LLM-KB) with COD prompting strategies to enhance English-to-Spanish biomedical translation. Specifically, MedCOD incorporates multi-layered domain knowledge, combines structured prompting with lightweight fine-tuning, and systematically evaluates multiple open-source LLMs under different adaptation settings. Our results demonstrate that MedCOD substantially improves translation quality, clinical accuracy, and contextual integrity, enabling open-source models to rival or even surpass proprietary systems.

Specifically, MedCOD differs from previous work in three key aspects:
1) it incorporates multi-layered domain knowledge, including translated medical terms, synonyms, and multilingual mappings from both UMLS and LLM-KB, to provide richer structured context during translation;
2) it combines structured prompting with lightweight fine-tuning, allowing models to better adapt to specialized biomedical content; and
3) it systematically evaluates multiple open-source LLMs under different adaptation settings, highlighting that domain-specific enrichment can enable open models to achieve or even surpass proprietary systems’ performance.
Figure~\ref{fig:medcod} illustrates how MedCOD systematically improves clinical translation quality by aligning with medical semantics and grammatical correctness across multiple translation challenges. Using established evaluation metrics such as SacreBLEU, ChrF++, and COMET, we demonstrate that integrating structured medical context through MedCOD significantly enhances translation quality.

Our experiments~\footnote{The source code is released at: \url{https://github.com/shahidul034/NoteAid-translation-EngToSpa} with CC-BY-NC 4.0 license.} show that applying MedCOD with fine-tuning yields substantial performance improvements across various open-source models. For instance, Phi-4 (14B) improves from a baseline BLEU score of 24.47 to 44.23 after MedCOD prompting and fine-tuning. Similarly, Qwen2.5-14B achieves 41.95 BLEU, and other models like Meta-Llama-3.1-8B also show notable gains. These results underscore the potential of combining domain-specific knowledge with open-source LLMs to advance high-quality, clinically meaningful medical translation.

\section{Methods}

\subsection{Overview}
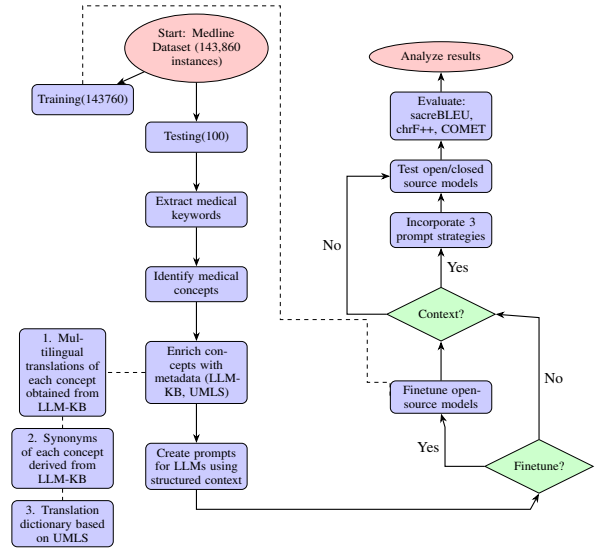
\begin{figure}
\centering
\scalebox{0.52}{
\begin{tikzpicture}[node distance=1cm and 1cm, auto]

    \tikzstyle{startstop} = [ellipse, rounded corners, minimum height=2em, minimum width=4em, 
                             draw, fill=red!20, text centered, font=\footnotesize, text width=6em]
    \tikzstyle{process} = [rectangle, minimum height=2em, minimum width=5em, rounded corners, 
                           draw, fill=blue!20, text centered, font=\footnotesize, text width=6em]
    \tikzstyle{decision} = [diamond, aspect=2, draw, fill=green!20, text centered, 
                            font=\footnotesize, text width=4em]
    \tikzstyle{arrow} = [thick, -Stealth, >=stealth]

    \node [startstop] (start) 
        {Start: Medline Dataset (143,860 instances)};
    \node [process, below = of start] (testing) 
        {Testing(100)};
    \node [process, below=of testing] (extract) 
        {Extract medical keywords};
    \node [process, below left=0.3cm and .3cm of start] (training) 
        {Training(143760)};
    \node [process, below=of extract] (identify) 
        {Identify medical concepts};
    \node [process, below=of identify] (enrich) 
        {Enrich concepts with metadata (LLM-KB, UMLS)};
    \node [process, below=of enrich] (prompts) 
        {Create prompts for LLMs using structured context};

    \node [process, left=1cm of enrich, text width=5em] (context1) 
        {1. Multilingual translations of each concept obtained from LLM-KB};
    \node [process, below=0.3cm of context1] (context2) 
        {2. Synonyms of each concept derived from LLM-KB};
    \node [process, below=0.3cm of context2] (context3) 
        {3. Translation dictionary based on UMLS};

    \node [decision, right=6cm of prompts] (finetune) 
        {Finetune?};

    \node [process, above left=1cm and 0.5cm of finetune, text width=6em] (finetune_model) 
        {Finetune open-source models};

    \node [decision, above=1cm of finetune_model] (context_yes) 
        {Context?};
    \node [process, above=1cm of context_yes] (inc) 
        {Incorporate 3 prompt strategies};
    \node [process, above=0.5cm of inc] (model) 
        {Test open/closed source models};
    \node [process, above=0.5cm of model] (eval) 
        {Evaluate: sacreBLEU, chrF++, COMET};
    \node [startstop, above=0.5cm of eval] (end) 
        {Analyze results};

    \draw [arrow] (start) -- (testing);
    \draw [arrow] (start) -- (training);
    \draw [arrow] (testing) -- (extract);
    \draw [arrow] (extract) -- (identify);
    \draw [arrow] (identify) -- (enrich);
    \draw [arrow] (enrich) -- (prompts);

    \draw [dashed] (enrich.west) |- (context1.east);
    \draw [dashed] (context1) -- (context2);
    \draw [dashed] (context2) -- (context3);
    

    \draw [arrow] (prompts.south) -- ++(0, -0.5) 
                  -- ++(8.45, 0) 
                  -- (finetune.south);

    \draw [arrow] (finetune.west)
        -- ++(-1,0) 
        -- ++(0,1) node[midway, left]{Yes}
        -- (finetune_model.south);

    \draw [arrow] (finetune.north)
        -- ++(0,3.1) node[midway, right]{No}
        -- (context_yes.east);

    \draw [arrow] (context_yes.north) -- node[midway, right]{Yes} (inc.south);

    \draw [arrow] (finetune_model.north) -- (context_yes.south);
    \draw [arrow] (inc.north) -- (model.south);
    \draw [arrow] (model.north) -- (eval.south);
    \draw [arrow] (eval.north) -- (end.south);

    \draw [arrow] (context_yes.west) -- ++(-1.0,0) |- (model.west) node[pos=.25, left]{No};

    \draw [dashed] (training.north)
        -- ++(0, 2)
        -- ++(5, 0)
        -- ++(0, -8)
        -- ++(2.2, 0)
        -- ++(0, -1.6)
        -| (finetune_model.west);
        
\end{tikzpicture}
}
\caption{Flowchart for illustrating the process of medical dataset preprocessing, structured prompt creation, and model evaluation for fine-tuning.}
\label{fig:flowchart}
\end{figure}

We evaluated open-source LLMs for their effectiveness in translating medical text from English to Spanish, comparing their performance across different strategies. 
Figure \ref{fig:flowchart} provides an overview of our framework, which involves the Medline dataset, various LLMs, and prompting methods.
Our analysis investigates the impact of different prompting styles based on our framework to identify the most suitable augmented knowledge (referred to as contextual information in this paper) for translation.
Additionally, we assess the effectiveness of fine-tuning techniques in enhancing model performance.

\subsection{Data source}
\label{sec:dataset_methods}
The ESPACMedlinePlus dataset is derived from the NIH’s MedlinePlus website, which provides medical articles on various health topics \cite{medline_dataset}. Most English articles on the site have corresponding human-translated Spanish versions. Following \cite{liu2015translating}, we utilized a collection of 2,999 such translated articles to construct a parallel-aligned corpus.

After data cleaning and sentence alignment, we obtained a training set of 143,760 sentences. To optimize testing, domain experts manually selected 100 instances from the Medline dataset, ensuring a balanced distribution of sentence lengths for robust evaluation. Although MedlinePlus content originates from medical articles on various health topics, it has been demonstrated that its language and terminology are sufficiently representative of electronic health record (EHR) notes, making it a practical proxy for clinical text in translation tasks. The test set size was kept small due to the substantial computational costs of generating three types of prompting strategies for each sentence using LLM-KB and UMLS, a process that is both resource-intensive and time-consuming.

\begin{table}[!ht]
\centering
\caption{Prompt Structures for Contextual Information}
\label{tab:prompt_single_column}
\scriptsize
\resizebox{\linewidth}{!}{%
\begin{tabular}{p{8cm}}
\hline
\textbf{Prompt Structure} \\ \hline

\textbf{Multilingual translations of each concept obtained from LLM-KB} \\
\quad - \textless{}Concept X\textgreater{}: \textless{}Auxiliary language 1\textgreater{}: \textless{}Word X in auxiliary-language 1\textgreater{}, 
\quad\quad \textless{}Auxiliary language 2\textgreater{}: \textless{}Word X in auxiliary-language 2\textgreater{}. \\
\quad - \textless{}Concept Y\textgreater{}: \textless{}Auxiliary language 1\textgreater{}: \textless{}Word Y in auxiliary-language 1\textgreater{}, 
\quad\quad \textless{}Auxiliary language 2\textgreater{}: \textless{}Word Y in auxiliary-language 2\textgreater{}. \\
\hline

\textbf{Synonyms of each concept derived from LLM-KB} \\
\quad Synonyms of \textless{}Concept name\textgreater{} in different languages: 
\textless{}Auxiliary language 1\textgreater{}: {[}synonym1, synonym2, ...{]}; 
\textless{}Auxiliary language 2\textgreater{}: {[}synonym1, ...{]}. \\
\hline

\textbf{Translation dictionary based on UMLS} \\
\quad \textless{}word X in source-language\textgreater{} means \textless{}word X in target-language\textgreater{}. \\
\quad \textless{}word X in auxiliary-language 1\textgreater{} means \textless{}word X in auxiliary-language 2\textgreater{}. \\
\hline

\end{tabular}
}
\end{table}

\subsection{UMLS- and LLM-KB-Enriched Medical Chain-of-Dictionary Framework} 
We developed two types of prompts: general prompts and structured prompts. 
General prompts were mainly used for direct translation without contextual information about the sentence, 
while structured prompts provided additional information to the models, enhancing sentence coherence and improving translation accuracy. 
The overall pipeline is illustrated in Figure~\ref{fig:flowchart}.

In our implementation, \textbf{LLM-KB} (Large Language Model as Knowledge Base) refers to using a pre-trained large language model (GPT-4o-mini) to retrieve structured medical knowledge, including: 
(1) multilingual translations of medical terms, and 
(2) synonyms of each medical concept across languages. 
These outputs form a lightweight knowledge base used to construct structured prompts for downstream translation tasks. 
Specifically, LLM-KB is queried with templated prompts to extract relevant information for each identified medical concept in the input, 
leveraging the model’s implicit medical and multilingual knowledge without requiring access to external structured databases beyond UMLS.

Following this approach, we first extracted medical keywords from sentences using LLM-KB, treating them as medical concepts for prompt development. 
To create effective prompts, we used both LLM-KB and UMLS as sources. 
For each medical concept in our testing set, we gathered various types of information, including synonyms from UMLS, synonyms derived from LLM-KB, and multilingual translations obtained from LLM-KB. 
We then designed three different structured prompt formats incorporating enriched medical concept metadata, enabling the model to better understand sentence meaning and structure. 
These structured prompts were compared against general prompts to evaluate which types of structured information were most critical for accurate translation. 
Results, summarized in Table~\ref{tab:prompt_single_column}, show that the optimal prompt type varied across models.

\begin{figure*}
    \centering
    \includegraphics[width=\linewidth]{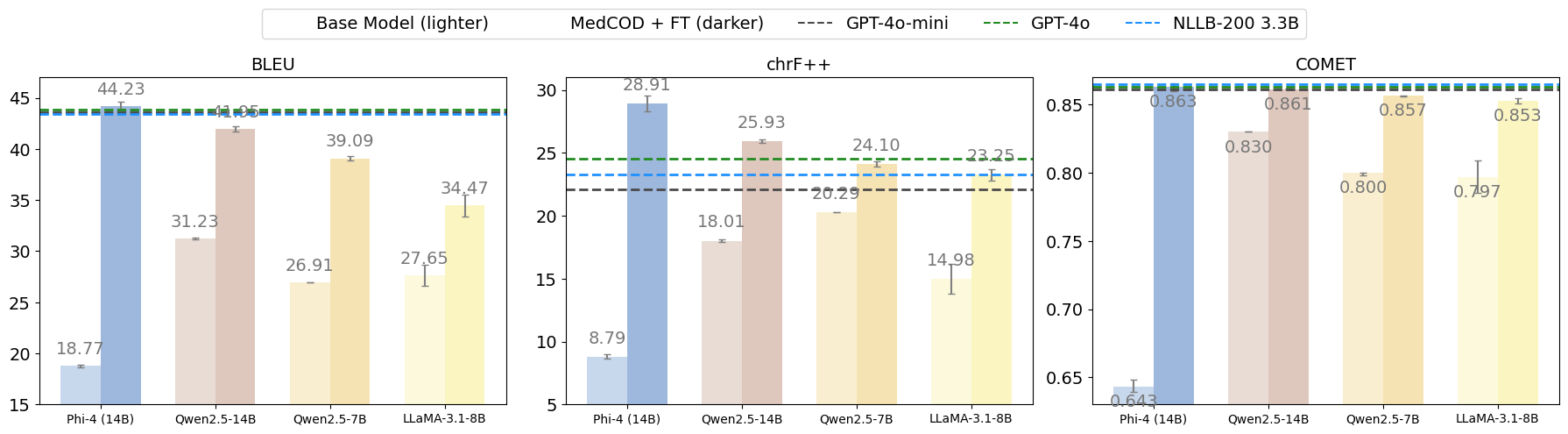}
    \caption{Performance comparison of four open-source LLMs (Phi-4, Qwen2.5-14B, Qwen2.5-7B, and LLaMA-3.1-8B) on ESPACMedlinePlus under two settings: base model (light) and MedCOD-enhanced fine-tuning (dark). Evaluation is conducted using BLEU, chrF++, and COMET metrics. Error bars indicate 95\% confidence intervals. The horizontal dashed line represents the performance of GPT-4o-mini. Results show that MedCOD + FT consistently improves model performance across all metrics, with Phi-4 (14B) achieving the best results overall. Note: The COMET score of the Phi-4 (14B) base model is too low (0.643 [0.639, 0.648]) and falls below the visible y-axis range, so it is not shown in the figure.}
    \label{fig:main_results}
\end{figure*}

\subsection{Fine-tuning with Low-Rank Adaptation (LoRA) and hyperparameter tuning} 
As shown in Figure~\ref{fig:flowchart}, the MedCOD framework can function purely as a prompting strategy to provide external knowledge to LLMs. 
In addition, for open-source LLMs, we can further enhance performance by fine-tuning the model to better utilize the provided contextual information.
Specifically, we employ LoRA \cite{lora}, a popular and lightweight fine-tuning technique that significantly reduces the number of trainable parameters. 
LoRA achieves this by injecting a small set of trainable weights into the model while keeping the original parameters frozen. 
This approach enables faster training, increased memory efficiency, and results in compact model weight files (only a few hundred megabytes), making them easier to store and share.

\subsection{Experimental Setup}

\subsubsection{Models} 

We evaluated open-source LLMs, including Phi-4 (14B) \cite{phi-4}, Qwen2.5-14B \cite{qwen2.5-14B}, Qwen2.5-7B \cite{qwen2.5-14B}, Meta-LLaMA-3.1-8B \cite{llama_model}, and used GPT-4o \cite{gpt4o}, GPT-4o Mini \cite{gpt4omini}, and NLLB-200 3.3B \cite{nllb} as baseline models. More detailed explanations of those models are given in the appendix.

\subsubsection{Datasets}

As described in Section~\ref{sec:dataset_methods}, we use the ESPACMedlinePlus~\cite{medline_dataset}, a parallel English–Spanish medical corpus derived from NIH’s MedlinePlus, for in-depth ablation studies and qualitative analysis of MedCOD.

\paragraph{WMT24}
To assess MedCOD’s applicability to diverse language pairs and broader medical translation scenarios, we incorporated the WMT 2024 Biomedical test set~\cite{neves2024findings}, which provides paragraph-level medical translation tasks across six language pairs: Portuguese–English, Spanish–English, Russian–English, German–English, French–English, and Italian–English, in both directions. For each language direction, we sampled 50 paragraph-level examples, resulting in a total of 600 test instances.

\paragraph{MultiClinSum}
To further evaluate MedCOD’s applicability beyond translation, we utilized the MultiClinSum dataset~\cite{rodriguez2025multiclinsum}, which contains clinical case reports designed for multilingual summarization tasks across four languages: English, Spanish, French, and Portuguese.
The dataset comprises 3,396 full-text clinical cases in English, 3,406 in Spanish, 3,469 in French, and 3,442 in Portuguese. 
We also used the MultiClinSum large-scale training datasets for model training across four languages. 
Each dataset contains 25,902 full-text and summary pairs in English, Spanish, French, and Portuguese.

\subsubsection{Evaluation} 

For \textbf{Machine Translation}, we evaluated translations against Spanish references using three metrics: SacreBLEU~\cite{sacrebleu}, ChrF++~\cite{chrf++}, and COMET~\cite{comet}. 
SacreBLEU and ChrF++ measure surface similarity at the word and character levels, respectively, while COMET uses neural models fine-tuned on human judgments to assess semantic adequacy and fluency.
For \textbf{Multilingual Medical Text Summarization}, we evaluated summarization performance using two widely adopted metrics: ROUGE~\cite{rouge} and BERTScore~\cite{bertscore}.
ROUGE measures sentence-level overlap between generated and reference summaries by focusing on the longest common subsequence. It is well-suited for extractive summarization.
BERTScore evaluates semantic similarity between generated and reference summaries using contextual embeddings from pre-trained BERT models, capturing meaning beyond exact token matches. It reports Precision, Recall, and F1, offering a balanced assessment of summary quality.
Complete metric definitions and formulas are provided in Appendix~\ref{appendix:summarization_metrics}.


\section{Main Results}

\paragraph{ESPACMedlinePlus} Figure~\ref{fig:main_results} presents the overall performance comparison of four open-source LLMs (e.g., Phi-4 (14B), Qwen2.5-14B, Qwen2.5-7B, and LLaMA-3.1-8B) under base and MedCOD-enhanced fine-tuning conditions. 
All models were evaluated using BLEU, chrF++, and COMET, with GPT-4o-mini/GPT-4o/nllb-200-3.3B as the baseline references (BLEU = 43.57/43.89/43.43, chrF++ = 22.13/24.52/23.25, COMET = 0.8615/0.8628/0.8648). 
MedCOD + FT consistently improves model performance across all metrics and models. 
Notably, Phi-4 (14B) with MedCOD + FT achieves the best results overall: BLEU = 44.23 (95\% CI: 43.82–44.64), chrF++ = 28.91 (28.29–29.53), and COMET = 0.863 (0.860–0.866), all exceeding GPT-4o-mini and GPT-4o. 
Other models also show substantial gains. 
For example, Qwen2.5-14B improves from BLEU = 31.23 (base) to 41.95, and chrF++ from 18.01 to 25.93; 
LLaMA-3.1-8B increases from BLEU = 27.65 to 34.47 and chrF++ from 14.98 to 23.25.
Importantly, all fine-tuned models surpass GPT-4o-mini in chrF++, indicating enhanced character-level translation quality.

\paragraph{WMT24} 
Appendix Table~\ref{tab:WMT24_resutls} summarizes MedCOD’s extension to paragraph-level medical translation in WMT24, covering 12 directions across six language pairs (50 high-complexity clinical paragraphs per direction). Sentence length analysis (Table~\ref{tab:WMT24_length}) highlights the substantial output lengths (up to 2,124 tokens) and supports context length configuration. Without fine-tuning, MedCOD-style contextual augmentation yields consistent gains in BLEU/chrF++ scores, particularly for morphologically rich or low-alignment pairs (e.g., en→ru, en→de). With fine-tuning, most directions further improve, with the largest boosts in BLEU for de→en (+9.29) and es→en (+11.21) over non-fine-tuned baselines. These results demonstrate that MedCOD’s benefits generalize beyond English–Spanish to multiple high-value medical translation settings.

\paragraph{MultiClinSum}
Appendix Table~\ref{tab:MultiClinSum_resutls} presents the overall performance on MultiClinSum. Across all four languages in MultiClinSUM~\cite{rodriguez2025multiclinsum}, Qwen2.5-14B-Instruct achieved its strongest performance when equipped with both MedCOD fine-tuning and contextual augmentation, with the most pronounced gains in recall-oriented metrics such as ROUGE-R and BERTScore-R (e.g., 0.8136 for English, 0.8100 for Spanish). While improvements were largest for high-resource pairs, consistent benefits were also observed in French and Portuguese, indicating MedCOD’s ability to recover salient medical information across languages. Removing either fine-tuning or contextual augmentation substantially degraded performance, especially in BERTScore-F and ROUGE-L, underscoring their complementary contributions. Together with our multilingual translation experiments, these results reinforce MedCOD’s robustness and generalizability across (1) different tasks (e.g., translation and summarization), (2) diverse languages, and (3) long, high-stakes medical texts.

\section{Ablation Study}

\begin{table}[!ht]
\centering
\resizebox{\linewidth}{!}{%
\begin{tabular}{|l|ccc|}
\hline
\textbf{Model} & BLEU & chrF++ & COMET \\
\hline
\multicolumn{4}{|l|}{\textbf{Block 1: w/o MedCOD, w/o FT}} \\
\hline
Phi-4 (14B)        & 18.77 (18.65, 18.89) & 8.79 (8.61, 8.97) & 0.643 (0.639, 0.648) \\
Qwen2.5-14B        & 31.23 (31.14, 31.32) & 18.01 (17.91, 18.12) & 0.830 (0.830, 0.830) \\
Qwen2.5-7B         & 26.91 (26.91, 26.91) & 20.29 (20.29, 20.29) & 0.800 (0.798, 0.800) \\
LLaMA-3.1-8B       & 27.65 (26.63, 28.67) & 14.98 (13.78, 16.18) & 0.797 (0.785, 0.809) \\
\hline
\multicolumn{4}{|l|}{\textbf{Block 2: w/ MedCOD, w/o FT} \small{(An asterisk (*) indicates a significant difference from Block 1.)}} \\
\hline
Phi-4 (14B)        & 32.71* (32.18, 33.24) & 21.86* (21.41, 22.31) & 0.819* (0.818, 0.820) \\
Qwen2.5-14B        & 35.55* (35.17, 36.58) & 22.09* (21.98, 22.68) & 0.848* (.832, .876) \\
Qwen2.5-7B         & 33.40* (33.40, 33.40) & 20.35* (20.35, 20.35) & 0.845* (0.845, 0.845) \\
LLaMA-3.1-8B       & 30.22* (29.23, 31.22) & 20.07* (19.26, 20.88) & 0.798 (0.788, 0.808) \\
\hline
\multicolumn{4}{|l|}{\textbf{Block 3: w/o MedCOD, w/ FT} \small(An asterisk (*) indicates a significant difference from Block 1.)} \\
\hline
Phi-4 (14B)        & 42.52* (42.11, 42.93) & 28.35* (27.73, 28.97) & 0.862* (0.859, 0.864) \\
Qwen2.5-14B        & 38.63* (38.52, 38.74) & 23.53* (23.36, 23.70) & 0.849* (0.849, 0.849) \\
Qwen2.5-7B         & \textbf{39.09*} (38.87, 39.31) & 23.94* (23.87, 24.01) & 0.856* (0.855, 0.856) \\
LLaMA-3.1-8B       & 34.47* (33.39, 35.55) & 22.11* (21.69, 22.53) & 0.850* (0.849, 0.852) \\
\hline
\multicolumn{4}{|l|}{\textbf{Block 4: w/ MedCOD, w/ FT} \small {(An asterisk (*) indicates a significant difference from Block 3.)}}\\
\hline
Phi-4 (14B)        & \textbf{44.23*} (43.82, 44.64) & \textbf{28.91} (28.29, 29.53) & \textbf{0.863} (0.860, 0.866) \\
Qwen2.5-14B        & \textbf{41.95*} (41.69, 42.21) & \textbf{25.93*} (25.81, 26.05) & \textbf{0.861*} (0.861, 0.862) \\
Qwen2.5-7B         & \textbf{39.09*} (38.87, 39.31) & \textbf{24.10} (23.90, 24.30) & \textbf{0.857*} (0.856, 0.857) \\
LLaMA-3.1-8B       & \textbf{34.47} (33.39, 35.55) & \textbf{23.25*} (22.79, 23.71) & \textbf{0.853*} (0.851, 0.855) \\
\hline
\end{tabular}
}
\caption{Performance of LLMs (with or without MedCOD and finetuning (FT) Across Contextual and Finetuning Settings (95\% CI) on ESPACMedlinePlus. An asterisk (*) indicates a statistically significant difference compared to the baseline condition described at the start of each block (p $<$ 0.05). Bold values represent the best-performing configuration per model.}
\label{tab:main_results}
\end{table}

\begin{table*}[!ht]
\centering
\resizebox{\textwidth}{!}{%
\begin{tabular}{|l|ccc|ccc|ccc|ccc|}
\hline
\multirow{2}{*}{\textbf{Model}} 
& \multicolumn{3}{c|}{\textbf{LLM-KB-Multilingual}} 
& \multicolumn{3}{c|}{\textbf{LLM-KB-Synonyms}} 
& \multicolumn{3}{c|}{\textbf{UMLS-Dict}} 
& \multicolumn{3}{c|}{\textbf{Direct Translation}} \\
\cline{2-13}
& BLEU & chrF++ & COMET & BLEU & chrF++ & COMET & BLEU & chrF++ & COMET & BLEU & chrF++ & COMET \\
\hline
Phi-4 14B (FT) 
& \textbf{44.23} & 28.91 & \textbf{0.8630}
& 42.10 & \textbf{29.47} & 0.8627
& 41.38 & 25.39 & 0.8541
& 42.52 & 28.35 & 0.8616 \\
\hline
Phi-4 14B (No FT)
& 24.47 & 12.89 & 0.7900
& 32.71 & \textbf{21.86} & 0.8190
& \textbf{35.89} & 20.17 & \textbf{0.8350}
& 18.77 & 8.79 & 0.6430 \\
\hline
Qwen2.5 14B (FT)
& \textbf{41.95} & 25.93 & \textbf{0.8614}
& 39.17 & 23.82 & 0.8572
& 39.47 & \textbf{26.28} & 0.8511
& 38.63 & 23.53 & 0.8491 \\
\hline
Qwen2.5 14B (No FT)
& \textbf{35.55} & \textbf{22.08} & \textbf{0.8480}
& 33.05 & 21.60 & 0.8420
& 34.18 & 21.77 & 0.8399
& 31.23 & 18.01 & 0.8300 \\
\hline
LLaMA-3.1-8B (FT)
& 33.15 & 19.67 & 0.8481
& \textbf{34.47} & \textbf{23.25} & \textbf{0.8526}
& \textbf{34.47} & 21.10 & 0.8483
& 33.12 & 22.11 & 0.8502 \\
\hline
LLaMA-3.1-8B (No FT)
& 28.64 & \textbf{21.84} & \textbf{0.8083}
& \textbf{30.22} & 20.07 & 0.7980
& 27.54 & 19.23 & 0.7921
& 27.65 & 14.98 & 0.7970 \\
\hline
Qwen2.5 7B (FT)
& 38.86 & 22.80 & 0.8565
& 38.43 & 24.05 & \textbf{0.8580}
& \textbf{39.09} & \textbf{24.10} & 0.8565
& 38.86 & 23.94 & 0.8555 \\
\hline
Qwen2.5 7B (No FT)
& \textbf{33.40} & \textbf{20.35} & \textbf{0.8451}
& 31.07 & 19.14 & 0.8211
& 24.50 & 16.58 & 0.7607
& 26.91 & 20.29 & 0.8000 \\
\hline
\end{tabular}
}
\caption{Mean Scores for Different Models Grouped by Context Type on ESPACMedlinePlus. Bold indicates the best performance prompts obtained by each LLM under MedCOD + (FT or No FT) settings. LLM-KB-Multilingual: Multilingual translations of each concept obtained from LLM-KB; LLM-KB-Synonyms: Synonyms of each concept derived from LLM-KB; UMLS-Dict: Translation dictionary based on UMLS.}
\label{tab:effectiveness_different_prompting}
\end{table*}

Due to space limitations, our ablation study mainly revolves around ESPACMedlinePlus.

\paragraph{Contributions of MedCOD Prompting and Fine-Tuning}
To understand how MedCOD contributes to performance improvements, Table~\ref{tab:main_results} presents a block-wise ablation study under four settings: (1) base model without MedCOD or fine-tuning, (2) MedCOD only, (3) fine-tuning only, and (4) MedCOD + FT. Comparing Block 1 and Block 2, contextual augmentation alone significantly enhances performance even without model fine-tuning. For instance, Phi-4 (14B) improves from BLEU = 18.77 to 32.71, and Qwen2.5-14B from BLEU = 31.23 to 35.55. These results demonstrate that structured prompts derived from UMLS and LLM-KB effectively enrich domain-specific understanding. 
In addition, while BLEU, chrF++, and COMET generally yield consistent evaluations, we observe occasional divergences that reflect their distinct emphases. For example, Qwen2.5-7B (Block 1: w/o MedCOD, w/o FT) achieves relatively higher chrF++ (20.29) despite a lower COMET score (0.800), while Qwen2.5-14B (Block 2: w/ MedCOD, w/o FT) presents the opposite trend with chrF++ = 22.09 and a much stronger COMET = 0.848. This indicates that chrF++ favors outputs with higher character-level overlap, such as morphologically accurate translations, whereas COMET emphasizes semantic preservation and fluency even when surface forms differ. These metric differences highlight the necessity of multi-dimensional evaluation in high-stakes domains like medical translation.

From Block 1 to Block 3, we observe the impact of fine-tuning in the absence of MedCOD. Phi-4 (14B) improves from BLEU = 18.77 to 42.52, and COMET increases from 0.643 to 0.862. Likewise, Qwen2.5-14B improves from BLEU = 31.23 to 38.63, and LLaMA-3.1-8B increases from 27.65 to 34.47. These consistent gains confirm that fine-tuning itself is a strong performance enhancer even without external context.

Finally, comparing Block 3 and Block 4 reveals that MedCOD significantly enhances performance when applied to fine-tuned models. For example, Qwen2.5-14B improves from BLEU = 38.63 to 41.95, and COMET from 0.8491 to 0.8614. Similarly, Phi-4 (14B) experiences additional improvements, with BLEU increasing from 42.52 to 44.23 and COMET rising from 0.8616 to 0.8630. These gains, though smaller in absolute magnitude, are statistically significant and indicate that contextual augmentation complements fine-tuning, pushing model performance beyond what fine-tuning alone can achieve.

\paragraph{Prompting Strategies and Model-Specific Variations}
To further investigate the impact of different types of structured context, Table~\ref{tab:effectiveness_different_prompting} compares multilingual translations, synonym expansion, and UMLS-based translation dictionaries. Across both fine-tuned and non-fine-tuned settings, multilingual translation prompts generally yield the highest scores. For instance, in the Phi-4 (FT) setting, multilingual prompts yield a COMET of 0.86299 and a BLEU of 44.23, outperforming synonyms (COMET = 0.86270) and UMLS dictionaries (0.85409). Similarly, in Qwen2.5-14B (FT), multilingual prompts again yield the best BLEU (41.95) and COMET (0.8614) scores. 
However, no single prompt structure is universally best across all models. For example, Phi-4 (No FT) achieves the highest BLEU (35.89) and COMET (0.835) using UMLS dictionary prompts, while LLaMA-3.1-8B (FT) achieves slightly better chrF++ (23.25) and COMET (0.8526) using synonym-based prompts. These results suggest that all three prompt types consistently outperform direct translation across models; however, the optimal prompting strategy may depend on the model architecture and fine-tuning status. Overall, multilingual prompting emerges as the most robust choice, but synonym- and dictionary-based prompts remain valuable, particularly in resource-constrained or base-model settings. This highlights a promising direction for future work in adaptive prompt selection based on model behavior and task requirements.

\section{Discussion}

\paragraph{Overall Effectiveness of MedCOD}
Our study demonstrates that the integration of domain-specific structured knowledge via the MedCOD framework substantially enhances the medical translation capabilities of open-source LLMs. As shown in Figure~\ref{fig:main_results} and Table~\ref{tab:main_results}, combining MedCOD prompting with fine-tuning enables models such as Phi-4 (14B) and Qwen2.5-14B to not only surpass their base performance but also exceed proprietary systems like GPT-4o on key metrics such as SacreBLEU, chrF++ and COMET. These findings reinforce the central hypothesis that clinical translation quality can be improved through both external knowledge (via LLM-KB/UMLS) and task-specific adaptation (e.g., fine-tuning).

Beyond aggregate scores, qualitative analysis further supports this conclusion. In Table~\ref{tab:ablation_case_studies1}, we compare translation outputs from GPT-4o and MedCOD-enhanced Phi-4 (14B). While both capture the overall meaning, Phi-4 produces a more medically accurate and professionally phrased output, e.g., using “sistema inmunitario” instead of “sistema inmunológico,” and “dificultad para respirar” rather than the more colloquial “falta de aliento.” Although Phi-4 introduces a minor grammatical error (“la trasplante”), it maintains the clinical register more faithfully. This case demonstrates how MedCOD-equipped open-source models can rival and even surpass proprietary systems in terms of biomedical translation fidelity.

While the three evaluation metrics generally reflect similar performance trends, they emphasize different aspects of translation quality. SacreBLEU and chrF++ measure surface-level similarity, capturing lexical and structural overlap between prediction and reference. In contrast, COMET prioritizes semantic adequacy and fluency, aligning more closely with human assessments of meaning. This distinction explains occasional discrepancies between scores and underscores the importance of combining form- and meaning-based metrics to assess clinical translation quality comprehensively.


\paragraph{Contributions of Prompting and Fine-Tuning}
The ablation analysis further reveals the stepwise contribution of each component in MedCOD. Contextual prompting alone significantly boosts performance even in non-finetuned models (e.g., Phi-4: BLEU +74.3\%); similarly, fine-tuning independently improves translation accuracy (e.g., Qwen2.5-14B: BLEU +23.6\%). Notably, the joint application of MedCOD and fine-tuning consistently yields the best performance across all models. Table~\ref{tab:ablation_case_studies2} qualitatively illustrates these trends: Case A shows how fine-tuning restores syntactic fluency (“y obtener una muestra”); Case B highlights MedCOD’s ability to recover complete noun phrases and terminological precision (“las venas y las arterias”); Case C demonstrates the cumulative benefits of both strategies, reinforcing grammatical correctness (“transpuestos”) and domain-appropriate phrasing (“la transposición de los grandes vasos”). Together, these examples align with the quantitative gains in BLEU, chrF++, and COMET, confirming the complementary nature of structured prompting and model adaptation.

\paragraph{Prompting Strategies and Model-Specific Variations}
Table~\ref{tab:effectiveness_different_prompting} further explores the impact of different types of structured prompts. Multilingual translation prompts outperform synonym expansion and UMLS dictionaries in most settings, especially for fine-tuned models like Phi-4 (COMET = 0.8630) and Qwen2.5-14B (BLEU = 41.95, COMET = 0.8614). However, prompt effectiveness varies depending on the LLM architecture and fine-tuning status. For instance, UMLS dictionary prompts yield the best BLEU (35.89) in Phi-4 without FT, and synonyms perform best in LLaMA-3.1-8B with FT. This variability suggests that a one-size-fits-all prompting strategy may be suboptimal, and future work could explore model-aware prompt selection or automatic ensemble prompting.

\paragraph{Error Analysis}

Despite these substantial quantitative and qualitative gains, our detailed error analysis reveals several key limitations in both proprietary and MedCOD-enhanced open-source models. As shown in Table~\ref{error_analysis}, issues such as lexical inaccuracies, grammatical inconsistencies, stylistic awkwardness, and improper register still persist, albeit to varying degrees across models. These issues not only affect output quality but also point to critical directions for future refinement, especially in high-stakes biomedical translation.
For instance, GPT-4o occasionally uses informal or colloquial terms such as “\textcolor{red}{falta de aliento}” for “shortness of breath,” which may reduce medical clarity. It also fails to maintain formal tone (“\textcolor{blue}{Obtienes hierro}”) and sometimes omits articles required for grammatical correctness (“el espalda”). In contrast, Phi-4-MedCOD-FT generally adheres to domain-appropriate vocabulary and formality but is not immune to errors, e.g., the incorrect article in “\textcolor{orange}{la trasplante}” and over-literal phrasing like “\textcolor{purple}{esto la hace sentir incómoda}.” A particularly nuanced example occurs in medical terminology: GPT-4o uses “IRM” (a French-derived abbreviation for MRI), while Phi-4 correctly uses “resonancia magnética” but inaccurately refers to metastasis with “\textcolor{red}{extendido}” instead of the medically precise “diseminado.”
These examples demonstrate that while MedCOD substantially improves translation quality, there remains a need for more robust handling of grammatical gender, medical term disambiguation, and context-aware register control. Future work may benefit from integrating error-type-aware reward models, curated medical glossaries with formality tags, or reinforcement learning with fine-grained human feedback to further reduce errors.

\paragraph{Inference Efficiency Analysis}
To assess the efficiency of the MedCOD pipeline, we measured the average per-sentence processing time (averaged over 100 test cases on a single NVIDIA A100 GPU). Table~\ref{tab:time_breakdown} presents a breakdown across four main components. While structured prompting adds an overhead of approximately $1.43$ seconds before final translation, the major cost is from the final inference step when using large models with long prompts. This trade-off between translation quality and computational cost will be further explored in future optimization efforts (e.g., caching, lightweight term retrieval).

\begin{table}[h]
\centering
\resizebox{\linewidth}{!}{%
\begin{tabular}{l r}
\toprule
\textbf{Component} & \textbf{Avg. Time (sec)} \\
\midrule
Keyword Extraction & 0.8712 \\
Keyword Translation (per keyword) & 0.1038 \\
Quality Check for Translated Terms & 0.4537 \\
Final Sentence Translation (with Prompt) & 8.5765 \\
\midrule
\textbf{Total Average Time per Sentence} & \textbf{10.01} \\
\bottomrule
\end{tabular}
}
\caption{Average per-sentence processing time for the MedCOD pipeline.}
\vspace{-3mm}
\label{tab:time_breakdown}
\end{table}

\section{Related Works}

\paragraph{Machine Translation}
Machine translation has long been a core NLP task, evolving from rule-based and statistical systems~\cite{koehn2003,och2003,tillmann2004,chiang2007,galley2004} to neural approaches~\cite{costa2022no,yao2023benchmarking,hendy2023good,brown2020language} that significantly improved fluency and coherence. Beyond improving architectures, researchers have explored incorporating external lexical resources into MT. Earlier methods integrated bilingual dictionaries into NMT as hard constraints~\cite{hokamp2017lexically, post2018fast} or soft constraints~\cite{song2019code,dinu2019training,chen2021lexical}. For example, Zhang et al.~\cite{zhang2016bridging} leveraged dictionary information for rare word translation, while Arthur et al.~\cite{arthur2016incorporating} combined lexicons with attention mechanisms. More recently, Lu et al.~\cite{lu2023chain} introduced the Chain-of-Dictionary (COD) framework, showing that structured lexical knowledge can substantially enhance translation quality in specialized domains. In parallel, large language models (LLMs) such as GPT, Claude, and LLaMA have demonstrated strong zero- and few-shot performance in general multilingual translation, often surpassing traditional NMT systems~\cite{yao2023benchmarking,hendy2023good}.

\paragraph{Medical Translation}
Medical translation introduces unique challenges due to domain-specific terminology, complex syntax, and the high stakes of clinical accuracy~\cite{mc1, mc2, mc3, mc4}. 
General-purpose MT systems such as Google Translate work for simple texts but frequently fail on longer or ambiguous sentences~\cite{liu2015translating,zeng2010}. Earlier efforts~\cite{weng2019, chen2017, patil2014} attempted to address these issues by constructing medical corpora, fine-tuning on terminology datasets, and evaluating MT on materials such as EHRs, patient education documents, and public health texts. Riina et al.~\cite{riina2024evaluation} evaluated GPT-4o’s English-to-Spanish medical translation, finding high fluency but persistent clinical inaccuracies and context errors.

\paragraph{LLMs in Medical Translation}
Building on their success in general MT, LLMs have recently been explored in biomedical and clinical settings. State-of-the-art models such as GPT-4, Claude, and LLaMA exhibit strong capabilities in translation, domain adaptation, and knowledge integration~\cite{gpt4, claude, llama, achiam2023gpt,tran2024bioinstruct}. 
Prior work has shown that LLMs can achieve high-quality translation even without task-specific training~\cite{mt1, mt2}.
In healthcare, LLMs and related AI systems have been tested for tasks ranging from diagnostics~\cite{mcduff2023towards,tu2024towards,yang2025unveiling,yao2024medqa} to health communication~\cite{wang2023notechat,tran2025medreadctrl,yao2023readme}, yet they remain underutilized in routine practice~\cite{ai_underuse, yao2025survey}. 
Importantly, while LLM-based translations are often linguistically natural, existing studies~\cite{riina2024evaluation} indicate that they still struggle with biomedical context preservation and clinical accuracy.

Building upon these insights, our MedCOD framework leverages multi-layered knowledge from the UMLS and an LLM-KB. By combining UMLS concept relations, synonym expansions, and multilingual mappings with COD-style prompting strategies, MedCOD enriches the translation process with structured, context-sensitive medical information. We further incorporate lightweight domain-specific fine-tuning to adapt open-source LLMs to the biomedical domain. Our experiments demonstrate that this hybrid strategy significantly improves translation quality, enhances clinical accuracy, and preserves contextual integrity, enabling open-source models to achieve or even surpass the performance of proprietary systems.

\section{Conclusion}
MedCOD provides a scalable and effective framework for enhancing biomedical translation by leveraging structured domain knowledge and targeted fine-tuning. Our results show that open-source LLMs, when equipped with rich medical context via UMLS and LLM-KB, can rival and even outperform proprietary systems like GPT-4o in clinical translation accuracy. By combining prompting and lightweight adaptation, MedCOD offers a practical pathway to improve cross-lingual health communication for underrepresented populations.

\section{Limitations}

This study presents several limitations that suggest directions for future research.

Firstly, the translation dataset is derived exclusively from MedlinePlus articles, which, while medically accurate and publicly available, tend to follow standardized formatting and controlled language. This may not fully capture the linguistic diversity and complexity found in other clinical domains, such as discharge summaries, progress notes, or specialty-specific documentation.

Secondly, our work focuses solely on English-to-Spanish translation. While this language pair is highly relevant in the U.S. context, further studies are needed to evaluate the adaptability of MedCOD to other language pairs, particularly those with lower resource availability or significant morphological divergence from English.

Thirdly, while MedCOD integrates domain knowledge using UMLS and LLM-KB, these knowledge sources have inherent limitations. Certain emerging medical concepts, abbreviations, or context-dependent expressions may be missing or incompletely represented. This could restrict the completeness of the structured prompts in specific scenarios.

Finally, although the evaluation includes widely used automatic metrics (e.g., BLEU, chrF++, and COMET), each captures different aspects of translation quality and may not fully reflect downstream clinical usability. Expanding evaluation to task-specific settings, such as cross-lingual question answering or information extraction, could provide more application-aligned assessments.

\section*{Acknowledgments}

This material is the result of work supported with resources and the use of facilities at the Center for Healthcare Organization and Implementation Research, VA Bedford Health Care.

\bibliography{emnlp25}

\begin{thebibliography}{74}
\providecommand{\natexlab}[1]{#1}

\bibitem[{Abdin et~al.(2024)Abdin, Aneja, Behl, Bubeck, Eldan, Gunasekar, Harrison, Hewett, Javaheripi, Kauffmann, Lee, Lee, Li, Liu, Mendes, Nguyen, Price, de~Rosa, Saarikivi, Salim, Shah, Wang, Ward, Wu, Yu, Zhang, and Zhang}]{phi-4}
Marah Abdin, Jyoti Aneja, Harkirat Behl, Sébastien Bubeck, Ronen Eldan, Suriya Gunasekar, Michael Harrison, Russell~J. Hewett, Mojan Javaheripi, Piero Kauffmann, James~R. Lee, Yin~Tat Lee, Yuanzhi Li, Weishung Liu, Caio C.~T. Mendes, Anh Nguyen, Eric Price, Gustavo de~Rosa, Olli Saarikivi, and 8 others. 2024.
\newblock \href {https://arxiv.org/abs/2412.08905} {Phi-4 technical report}.
\newblock \emph{Preprint}, arXiv:2412.08905.

\bibitem[{Achiam et~al.(2023{\natexlab{a}})Achiam, Adler, Agarwal, Ahmad, Akkaya, Aleman, Almeida, Altenschmidt, Altman, Anadkat et~al.}]{gpt4}
Josh Achiam, Steven Adler, Sandhini Agarwal, Lama Ahmad, Ilge Akkaya, Florencia~Leoni Aleman, Diogo Almeida, Janko Altenschmidt, Sam Altman, Shyamal Anadkat, and 1 others. 2023{\natexlab{a}}.
\newblock Gpt-4 technical report.
\newblock \emph{arXiv preprint arXiv:2303.08774}.

\bibitem[{Achiam et~al.(2023{\natexlab{b}})Achiam, Adler, Agarwal, Ahmad, Akkaya, Aleman, Almeida, Altenschmidt, Altman, Anadkat et~al.}]{achiam2023gpt}
Josh Achiam, Steven Adler, Sandhini Agarwal, Lama Ahmad, Ilge Akkaya, Florencia~Leoni Aleman, Diogo Almeida, Janko Altenschmidt, Sam Altman, Shyamal Anadkat, and 1 others. 2023{\natexlab{b}}.
\newblock Gpt-4 technical report.
\newblock \emph{arXiv preprint arXiv:2303.08774}.

\bibitem[{Arthur et~al.(2016)Arthur, Neubig, and Nakamura}]{arthur2016incorporating}
Philip Arthur, Graham Neubig, and Satoshi Nakamura. 2016.
\newblock Incorporating discrete translation lexicons into neural machine translation.
\newblock \emph{arXiv preprint arXiv:1606.02006}.

\bibitem[{Brown et~al.(1990)Brown, Cocke, Della~Pietra, Della~Pietra, Jelinek, Lafferty, Mercer, and Roossin}]{brown1990}
Peter~F Brown, John Cocke, Stephen~A Della~Pietra, Vincent~J Della~Pietra, Frederick Jelinek, John Lafferty, Robert~L Mercer, and Paul~S Roossin. 1990.
\newblock A statistical approach to machine translation.
\newblock \emph{Computational linguistics}, 16(2):79--85.

\bibitem[{Brown et~al.(2020{\natexlab{a}})Brown, Mann, Ryder, Subbiah, Kaplan, Dhariwal, Neelakantan, Shyam, Sastry, Askell, Agarwal, Herbert-Voss, Krueger, Henighan, Child, Ramesh, Ziegler, Wu, Winter, Hesse, Chen, Sigler, Litwin, Gray, Chess, Clark, Berner, McCandlish, Radford, Sutskever, and Amodei}]{mt1}
Tom Brown, Benjamin Mann, Nick Ryder, Melanie Subbiah, Jared~D Kaplan, Prafulla Dhariwal, Arvind Neelakantan, Pranav Shyam, Girish Sastry, Amanda Askell, Sandhini Agarwal, Ariel Herbert-Voss, Gretchen Krueger, Tom Henighan, Rewon Child, Aditya Ramesh, Daniel Ziegler, Jeffrey Wu, Clemens Winter, and 12 others. 2020{\natexlab{a}}.
\newblock Language models are few-shot learners.
\newblock In \emph{Advances in Neural Information Processing Systems}, volume~33, pages 1877--1901. Curran Associates, Inc.

\bibitem[{Brown et~al.(2020{\natexlab{b}})Brown, Mann, Ryder, Subbiah, Kaplan, Dhariwal, Neelakantan, Shyam, Sastry, Askell et~al.}]{brown2020language}
Tom Brown, Benjamin Mann, Nick Ryder, Melanie Subbiah, Jared~D Kaplan, Prafulla Dhariwal, Arvind Neelakantan, Pranav Shyam, Girish Sastry, Amanda Askell, and 1 others. 2020{\natexlab{b}}.
\newblock Language models are few-shot learners.
\newblock \emph{Advances in neural information processing systems}, 33:1877--1901.

\bibitem[{Chen et~al.(2021)Chen, Chen, Wang, and Li}]{chen2021lexical}
Guanhua Chen, Yun Chen, Yong Wang, and Victor~OK Li. 2021.
\newblock Lexical-constraint-aware neural machine translation via data augmentation.
\newblock In \emph{Proceedings of the Twenty-Ninth International Conference on International Joint Conferences on Artificial Intelligence}, pages 3587--3593.

\bibitem[{Chen et~al.(2017)Chen, Acosta, Barry et~al.}]{chen2017}
Xuewei Chen, Sandra Acosta, Adam~E Barry, and 1 others. 2017.
\newblock Machine or human? evaluating the quality of a language translation mobile app for diabetes education material.
\newblock \emph{JMIR diabetes}, 2(1):e7446.

\bibitem[{Chiang(2007)}]{chiang2007}
David Chiang. 2007.
\newblock Hierarchical phrase-based translation.
\newblock \emph{computational linguistics}, 33(2):201--228.

\bibitem[{Costa-Juss{\`a} et~al.(2022)Costa-Juss{\`a}, Cross, {\c{C}}elebi, Elbayad, Heafield, Heffernan, Kalbassi, Lam, Licht, Maillard et~al.}]{costa2022no}
Marta~R Costa-Juss{\`a}, James Cross, Onur {\c{C}}elebi, Maha Elbayad, Kenneth Heafield, Kevin Heffernan, Elahe Kalbassi, Janice Lam, Daniel Licht, Jean Maillard, and 1 others. 2022.
\newblock No language left behind: Scaling human-centered machine translation.
\newblock \emph{arXiv preprint arXiv:2207.04672}.

\bibitem[{Daniel~Han and team(2023)}]{unsloth}
Michael~Han Daniel~Han and Unsloth team. 2023.
\newblock \href {http://github.com/unslothai/unsloth} {Unsloth}.

\bibitem[{Delbanco et~al.(2012)Delbanco, Walker, Bell, Darer, Elmore, Farag, Feldman, Mejilla, Ngo, Ralston et~al.}]{delbanco2012inviting}
Tom Delbanco, Jan Walker, Sigall~K Bell, Jonathan~D Darer, Joann~G Elmore, Nadine Farag, Henry~J Feldman, Roanne Mejilla, Long Ngo, James~D Ralston, and 1 others. 2012.
\newblock Inviting patients to read their doctors' notes: a quasi-experimental study and a look ahead.
\newblock \emph{Annals of internal medicine}, 157(7):461--470.

\bibitem[{Dew et~al.(2018)Dew, Turner, Choi, Bosold, and Kirchhoff}]{mc3}
Kristin~N. Dew, Anne~M. Turner, Yong~K. Choi, Alyssa Bosold, and Katrin Kirchhoff. 2018.
\newblock \href {https://doi.org/10.1016/j.jbi.2018.07.018} {Development of machine translation technology for assisting health communication: A systematic review}.
\newblock \emph{Journal of Biomedical Informatics}, 85:56--67.

\bibitem[{Dinu et~al.(2019)Dinu, Mathur, Federico, and Al-Onaizan}]{dinu2019training}
Georgiana Dinu, Prashant Mathur, Marcello Federico, and Yaser Al-Onaizan. 2019.
\newblock Training neural machine translation to apply terminology constraints.
\newblock \emph{arXiv preprint arXiv:1906.01105}.

\bibitem[{Du{\v{s}}ek et~al.(2014)Du{\v{s}}ek, Hajic, Hlav{\'a}{\v{c}}ov{\'a}, Nov{\'a}k, Pecina, Rosa, Tamchyna, Ure{\v{s}}ov{\'a}, and Zeman}]{mc2}
Ond{\v{r}}ej Du{\v{s}}ek, Jan Hajic, Jaroslava Hlav{\'a}{\v{c}}ov{\'a}, Michal Nov{\'a}k, Pavel Pecina, Rudolf Rosa, Ale{\v{s}} Tamchyna, Zdenka Ure{\v{s}}ov{\'a}, and Daniel Zeman. 2014.
\newblock Machine translation of medical texts in the khresmoi project.
\newblock In \emph{Proceedings of the Ninth Workshop on Statistical Machine Translation}, pages 221--228.

\bibitem[{Gabay(2017)}]{gabay201721st}
Michael Gabay. 2017.
\newblock 21st century cures act.
\newblock \emph{Hospital pharmacy}, 52(4):264--265.

\bibitem[{Galley et~al.(2004)Galley, Hopkins, Knight, and Marcu}]{galley2004}
Michel Galley, Mark Hopkins, Kevin Knight, and Daniel Marcu. 2004.
\newblock What’s in a translation rule?
\newblock In \emph{Proc. of HLT-NAACL}, pages 273--280.

\bibitem[{Grattafiori et~al.(2024)Grattafiori, Dubey, Jauhri, Pandey, Kadian, Al-Dahle, Letman, Mathur, Schelten, Vaughan, Yang, Fan, Goyal, Hartshorn, Yang, Mitra, Sravankumar, Korenev, Hinsvark, Rao, Zhang, Rodriguez, Gregerson, Spataru, Roziere, Biron, Tang, Chern, Caucheteux, Nayak, Bi, Marra, McConnell, Keller, Touret, Wu, Wong, Ferrer, Nikolaidis, Allonsius, Song, Pintz, Livshits, Wyatt, Esiobu, Choudhary, Mahajan, Garcia-Olano, Perino, Hupkes, Lakomkin, AlBadawy, Lobanova, Dinan, Smith, Radenovic, Guzmán, Zhang, Synnaeve, Lee, Anderson, Thattai, Nail, Mialon, Pang, Cucurell, Nguyen, Korevaar, Xu, Touvron, Zarov, Ibarra, Kloumann, Misra, Evtimov, Zhang, Copet, Lee, Geffert, Vranes, Park, Mahadeokar, Shah, van~der Linde, Billock, Hong, Lee, Fu, Chi, Huang, Liu, Wang, Yu, Bitton, Spisak, Park, Rocca, Johnstun, Saxe, Jia, Alwala, Prasad, Upasani, Plawiak, Li, Heafield, Stone, El-Arini, Iyer, Malik, Chiu, Bhalla, Lakhotia, Rantala-Yeary, van~der Maaten, Chen, Tan, Jenkins, Martin, Madaan, Malo, Blecher,
  Landzaat, de~Oliveira, Muzzi, Pasupuleti, Singh, Paluri, Kardas, Tsimpoukelli, Oldham, Rita, Pavlova, Kambadur, Lewis, Si, Singh, Hassan, Goyal, Torabi, Bashlykov, Bogoychev, Chatterji, Zhang, Duchenne, Çelebi, Alrassy, Zhang, Li, Vasic, Weng, Bhargava, Dubal, Krishnan, Koura, Xu, He, Dong, Srinivasan, Ganapathy, Calderer, Cabral, Stojnic, Raileanu, Maheswari, Girdhar, Patel, Sauvestre, Polidoro, Sumbaly, Taylor, Silva, Hou, Wang, Hosseini, Chennabasappa, Singh, Bell, Kim, Edunov, Nie, Narang, Raparthy, Shen, Wan, Bhosale, Zhang, Vandenhende, Batra, Whitman, Sootla, Collot, Gururangan, Borodinsky, Herman, Fowler, Sheasha, Georgiou, Scialom, Speckbacher, Mihaylov, Xiao, Karn, Goswami, Gupta, Ramanathan, Kerkez, Gonguet, Do, Vogeti, Albiero, Petrovic, Chu, Xiong, Fu, Meers, Martinet, Wang, Wang, Tan, Xia, Xie, Jia, Wang, Goldschlag, Gaur, Babaei, Wen, Song, Zhang, Li, Mao, Coudert, Yan, Chen, and Papakipos}]{llama_model}
Aaron Grattafiori, Abhimanyu Dubey, Abhinav Jauhri, Abhinav Pandey, Abhishek Kadian, Ahmad Al-Dahle, Aiesha Letman, Akhil Mathur, Alan Schelten, Alex Vaughan, Amy Yang, Angela Fan, Anirudh Goyal, Anthony Hartshorn, Aobo Yang, Archi Mitra, Archie Sravankumar, Artem Korenev, Arthur Hinsvark, and 217 others. 2024.
\newblock \href {https://arxiv.org/abs/2407.21783} {The llama 3 herd of models}.
\newblock \emph{Preprint}, arXiv:2407.21783.

\bibitem[{Haldar et~al.(2023)Haldar, Pillai, and Artiga}]{haldar2023overview}
Sweta Haldar, Drishti Pillai, and Samantha Artiga. 2023.
\newblock Overview of health coverage and care for individuals with limited english proficiency (lep).
\newblock \emph{KFF}.

\bibitem[{Hendy et~al.(2023)Hendy, Abdelrehim, Sharaf, Raunak, Gabr, Matsushita, Kim, Afify, and Awadalla}]{hendy2023good}
Amr Hendy, Mohamed Abdelrehim, Amr Sharaf, Vikas Raunak, Mohamed Gabr, Hitokazu Matsushita, Young~Jin Kim, Mohamed Afify, and Hany~Hassan Awadalla. 2023.
\newblock How good are gpt models at machine translation? a comprehensive evaluation.
\newblock \emph{arXiv preprint arXiv:2302.09210}.

\bibitem[{Hokamp and Liu(2017)}]{hokamp2017lexically}
Chris Hokamp and Qun Liu. 2017.
\newblock Lexically constrained decoding for sequence generation using grid beam search.
\newblock \emph{arXiv preprint arXiv:1704.07138}.

\bibitem[{Hu et~al.(2021)Hu, Shen, Wallis, Allen-Zhu, Li, Wang, Wang, and Chen}]{lora}
Edward~J. Hu, Yelong Shen, Phillip Wallis, Zeyuan Allen-Zhu, Yuanzhi Li, Shean Wang, Lu~Wang, and Weizhu Chen. 2021.
\newblock \href {https://arxiv.org/abs/2106.09685} {Lora: Low-rank adaptation of large language models}.
\newblock \emph{Preprint}, arXiv:2106.09685.

\bibitem[{Kapoor et~al.(2022)Kapoor, Corrales, Flores, Feng, and Cata}]{kapoor2022}
Ravish Kapoor, German Corrales, Manuel~P Flores, Lei Feng, and Juan~P Cata. 2022.
\newblock Use of neural machine translation software for patients with limited english proficiency to assess postoperative pain and nausea.
\newblock \emph{JAMA Network Open}, 5(3):e221485--e221485.

\bibitem[{Kayastha et~al.(2018)Kayastha, Pollak, and LeBlanc}]{kayastha2018open}
Neha Kayastha, Kathryn~I Pollak, and Thomas~W LeBlanc. 2018.
\newblock Open oncology notes: a qualitative study of oncology patients’ experiences reading their cancer care notes.
\newblock \emph{Journal of Oncology Practice}, 14(4):e251--e258.

\bibitem[{Koehn et~al.(2003)Koehn, Och, and Marcu}]{koehn2003}
Philipp Koehn, Franz~Josef Och, and Daniel Marcu. 2003.
\newblock Statistical phrase-based translation.
\newblock In \emph{2003 Conference of the North American Chapter of the Association for Computational Linguistics on Human Langauge Technology (HLT-NAACL 2003)}, pages 48--54. Association for Computational Linguistics.

\bibitem[{Lin(2004)}]{rouge}
Chin-Yew Lin. 2004.
\newblock \href {https://aclanthology.org/W04-1013/} {{ROUGE}: A package for automatic evaluation of summaries}.
\newblock In \emph{Text Summarization Branches Out}, pages 74--81, Barcelona, Spain. Association for Computational Linguistics.

\bibitem[{Lin et~al.(2022)Lin, Mihaylov, Artetxe, Wang, Chen, Simig, Ott, Goyal, Bhosale, Du, Pasunuru, Shleifer, Koura, Chaudhary, O{'}Horo, Wang, Zettlemoyer, Kozareva, Diab, Stoyanov, and Li}]{mt2}
Xi~Victoria Lin, Todor Mihaylov, Mikel Artetxe, Tianlu Wang, Shuohui Chen, Daniel Simig, Myle Ott, Naman Goyal, Shruti Bhosale, Jingfei Du, Ramakanth Pasunuru, Sam Shleifer, Punit~Singh Koura, Vishrav Chaudhary, Brian O{'}Horo, Jeff Wang, Luke Zettlemoyer, Zornitsa Kozareva, Mona Diab, and 2 others. 2022.
\newblock \href {https://doi.org/10.18653/v1/2022.emnlp-main.616} {Few-shot learning with multilingual generative language models}.
\newblock In \emph{Proceedings of the 2022 Conference on Empirical Methods in Natural Language Processing}, pages 9019--9052, Abu Dhabi, United Arab Emirates. Association for Computational Linguistics.

\bibitem[{Lindberg et~al.(1993)Lindberg, Humphreys, and McCray}]{lindberg1993unified}
Donald~AB Lindberg, Betsy~L Humphreys, and Alexa~T McCray. 1993.
\newblock The unified medical language system.
\newblock \emph{Yearbook of medical informatics}, 2(01):41--51.

\bibitem[{Liu and Cai(2015{\natexlab{a}})}]{liu2015translating}
Weisong Liu and Shu Cai. 2015{\natexlab{a}}.
\newblock Translating electronic health record notes from english to spanish: A preliminary study.
\newblock In \emph{Proceedings of BioNLP 15}, pages 134--140.

\bibitem[{Liu and Cai(2015{\natexlab{b}})}]{medline_dataset}
Weisong Liu and Shu Cai. 2015{\natexlab{b}}.
\newblock \href {https://doi.org/10.18653/v1/W15-3816} {Translating electronic health record notes from {E}nglish to {S}panish: A preliminary study}.
\newblock In \emph{Proceedings of {B}io{NLP} 15}, pages 134--140, Beijing, China. Association for Computational Linguistics.

\bibitem[{Liu et~al.(2024)Liu, Duan, Kim, Zhang, Jee, Maharjan, Huang, Du, and Jiang}]{claude}
Xu~Liu, Chaoli Duan, Min-kyu Kim, Lu~Zhang, Eunjin Jee, Beenu Maharjan, Yuwei Huang, Dan Du, and Xian Jiang. 2024.
\newblock Claude 3 opus and chatgpt with gpt-4 in dermoscopic image analysis for melanoma diagnosis: comparative performance analysis.
\newblock \emph{JMIR Medical Informatics}, 12:e59273.

\bibitem[{Lu et~al.(2023)Lu, Yang, Huang, Zhang, Lam, and Wei}]{lu2023chain}
Hongyuan Lu, Haoran Yang, Haoyang Huang, Dongdong Zhang, Wai Lam, and Furu Wei. 2023.
\newblock Chain-of-dictionary prompting elicits translation in large language models.
\newblock \emph{arXiv preprint arXiv:2305.06575}.

\bibitem[{Lu et~al.(2024)Lu, Yang, Huang, Zhang, Lam, and Wei}]{codpaper}
Hongyuan Lu, Haoran Yang, Haoyang Huang, Dongdong Zhang, Wai Lam, and Furu Wei. 2024.
\newblock Chain-of-dictionary prompting elicits translation in large language models.
\newblock In \emph{Proceedings of the 2024 Conference on Empirical Methods in Natural Language Processing}, pages 958--976.

\bibitem[{McDuff et~al.(2023)McDuff, Schaekermann, Tu, Palepu, Wang, Garrison, Singhal, Sharma, Azizi, Kulkarni et~al.}]{mcduff2023towards}
Daniel McDuff, Mike Schaekermann, Tao Tu, Anil Palepu, Amy Wang, Jake Garrison, Karan Singhal, Yash Sharma, Shekoofeh Azizi, Kavita Kulkarni, and 1 others. 2023.
\newblock Towards accurate differential diagnosis with large language models.
\newblock \emph{arXiv preprint arXiv:2312.00164}.

\bibitem[{McQuaid and Landier(2018)}]{mcquaid2018cultural}
Elizabeth~L McQuaid and Wendy Landier. 2018.
\newblock Cultural issues in medication adherence: disparities and directions.
\newblock \emph{Journal of general internal medicine}, 33(2):200--206.

\bibitem[{Mehandru et~al.(2022)Mehandru, Robertson, and Salehi}]{mc1}
Nikita Mehandru, Samantha Robertson, and Niloufar Salehi. 2022.
\newblock \href {https://doi.org/10.1145/3531146.3533244} {Reliable and safe use of machine translation in medical settings}.
\newblock In \emph{Proceedings of the 2022 ACM Conference on Fairness, Accountability, and Transparency}, FAccT '22, page 2016–2025, New York, NY, USA. Association for Computing Machinery.

\bibitem[{Neves et~al.(2024)Neves, Grozea, Thomas, Roller, Bawden, N{\'e}v{\'e}ol, Castle, Bonato, Du~Nunzio, Vezzani et~al.}]{neves2024findings}
Mariana Neves, Cristian Grozea, Philippe Thomas, Roland Roller, Rachel Bawden, Aur{\'e}lie N{\'e}v{\'e}ol, Steffen Castle, Vanessa Bonato, Giorgio~Maria Du~Nunzio, Federica Vezzani, and 1 others. 2024.
\newblock Findings of the wmt 2024 biomedical translation shared task: Test sets on abstract level.
\newblock In \emph{Proceedings of the Ninth Conference on Machine Translation}.

\bibitem[{Och(2003)}]{och2003}
Franz~Josef Och. 2003.
\newblock \emph{Statistical machine translation: From single word models to alignment templates}.
\newblock Ph.D. thesis, Aachen, Techn. Hochsch., Diss., 2002.

\bibitem[{{OpenAI}(2024{\natexlab{a}})}]{gpt4omini}
{OpenAI}. 2024{\natexlab{a}}.
\newblock \href {https://openai.com/index/gpt-4o-mini-advancing-cost-efficient-intelligence/} {Gpt-4o mini: Advancing cost-efficient intelligence}.

\bibitem[{{OpenAI}(2024{\natexlab{b}})}]{gpt4o}
{OpenAI}. 2024{\natexlab{b}}.
\newblock \href {https://openai.com/index/hello-gpt-4o/} {Hello gpt-4o}.

\bibitem[{Pagallo et~al.(2024)Pagallo, O’Sullivan, Nevejans, Holzinger, Friebe, Jeanquartier, Jean-Quartier, and Miernik}]{ai_underuse}
Ugo Pagallo, Shane O’Sullivan, Nathalie Nevejans, Andreas Holzinger, Michael Friebe, Fleur Jeanquartier, Claire Jean-Quartier, and Arkadiusz Miernik. 2024.
\newblock The underuse of ai in the health sector: Opportunity costs, success stories, risks and recommendations.
\newblock \emph{Health and Technology}, 14(1):1--14.

\bibitem[{Patil and Davies(2014)}]{patil2014}
Sumant Patil and Patrick Davies. 2014.
\newblock Use of google translate in medical communication: evaluation of accuracy.
\newblock \emph{Bmj}, 349.

\bibitem[{Popovi{\'c}(2017)}]{chrf++}
Maja Popovi{\'c}. 2017.
\newblock \href {https://doi.org/10.18653/v1/W17-4770} {chr{F}++: words helping character n-grams}.
\newblock In \emph{Proceedings of the Second Conference on Machine Translation}, pages 612--618, Copenhagen, Denmark. Association for Computational Linguistics.

\bibitem[{Post(2018)}]{sacrebleu}
Matt Post. 2018.
\newblock \href {https://www.aclweb.org/anthology/W18-6319} {A call for clarity in reporting {BLEU} scores}.
\newblock In \emph{Proceedings of the Third Conference on Machine Translation: Research Papers}, pages 186--191, Belgium, Brussels. Association for Computational Linguistics.

\bibitem[{Post and Vilar(2018)}]{post2018fast}
Matt Post and David Vilar. 2018.
\newblock Fast lexically constrained decoding with dynamic beam allocation for neural machine translation.
\newblock \emph{arXiv preprint arXiv:1804.06609}.

\bibitem[{Rei et~al.(2022)Rei, C.~de Souza, Alves, Zerva, Farinha, Glushkova, Lavie, Coheur, and Martins}]{comet}
Ricardo Rei, Jos{\'e}~G. C.~de Souza, Duarte Alves, Chrysoula Zerva, Ana~C Farinha, Taisiya Glushkova, Alon Lavie, Luisa Coheur, and Andr{\'e} F.~T. Martins. 2022.
\newblock \href {https://aclanthology.org/2022.wmt-1.52} {{COMET}-22: Unbabel-{IST} 2022 submission for the metrics shared task}.
\newblock In \emph{Proceedings of the Seventh Conference on Machine Translation (WMT)}, pages 578--585, Abu Dhabi, United Arab Emirates (Hybrid). Association for Computational Linguistics.

\bibitem[{Riina et~al.(2024)Riina, Patlolla, Joya, Bautista, Olivar-Villanueva, and Kumar}]{riina2024evaluation}
Nicholas Riina, Likhitha Patlolla, Camilo~Hernandez Joya, Roger Bautista, Melissa Olivar-Villanueva, and Anish Kumar. 2024.
\newblock An evaluation of english to spanish medical translation by large language models.
\newblock In \emph{Proceedings of the 16th Conference of the Association for Machine Translation in the Americas (Volume 2: User Track)}, pages 222--236.

\bibitem[{Rodríguez-Ortega et~al.(2025)Rodríguez-Ortega, Rodríguez-Lopez, Lima-López, Escolano, Melero, Pratesi, Vigil-Giménez, Fernandez, Farré-Maduell, and Krallinger}]{rodriguez2025multiclinsum}
María Rodríguez-Ortega, Eduardo Rodríguez-Lopez, Sandra Lima-López, Carlos Escolano, Marta Melero, Lorenzo Pratesi, Laura Vigil-Giménez, Lucía Fernandez, Eva Farré-Maduell, and Martin Krallinger. 2025.
\newblock Overview of multiclinsum task at bioasq 2025: Evaluation of clinical case summarization strategies for multiple languages: Data, evaluation, resources and results.
\newblock In \emph{CLEF 2025 Working Notes}. CEUR Workshop Proceedings.
\newblock To appear.

\bibitem[{Root et~al.(2016)Root, Oster, L.~Jackson, Mejilla, Walker, and Elmore}]{root2016characteristics}
Joseph Root, Natalia~V Oster, Sara L.~Jackson, Roanne Mejilla, Jan Walker, and Joann~G Elmore. 2016.
\newblock Characteristics of patients who report confusion after reading their primary care clinic notes online.
\newblock \emph{Health communication}, 31(6):778--781.

\bibitem[{Song et~al.(2019)Song, Zhang, Yu, Luo, Wang, and Zhang}]{song2019code}
Kai Song, Yue Zhang, Heng Yu, Weihua Luo, Kun Wang, and Min Zhang. 2019.
\newblock Code-switching for enhancing nmt with pre-specified translation.
\newblock \emph{arXiv preprint arXiv:1904.09107}.

\bibitem[{Team et~al.(2022)Team, Costa-jussà, Cross, Çelebi, Elbayad, Heafield, Heffernan, Kalbassi, Lam, Licht, Maillard, Sun, Wang, Wenzek, Youngblood, Akula, Barrault, Gonzalez, Hansanti, Hoffman, Jarrett, Sadagopan, Rowe, Spruit, Tran, Andrews, Ayan, Bhosale, Edunov, Fan, Gao, Goswami, Guzmán, Koehn, Mourachko, Ropers, Saleem, Schwenk, and Wang}]{nllb}
NLLB Team, Marta~R. Costa-jussà, James Cross, Onur Çelebi, Maha Elbayad, Kenneth Heafield, Kevin Heffernan, Elahe Kalbassi, Janice Lam, Daniel Licht, Jean Maillard, Anna Sun, Skyler Wang, Guillaume Wenzek, Al~Youngblood, Bapi Akula, Loic Barrault, Gabriel~Mejia Gonzalez, Prangthip Hansanti, and 20 others. 2022.
\newblock \href {https://arxiv.org/abs/2207.04672} {No language left behind: Scaling human-centered machine translation}.
\newblock \emph{arXiv preprint arXiv:2207.04672}.

\bibitem[{Team(2024)}]{qwen2.5-14B}
Qwen Team. 2024.
\newblock \href {https://qwenlm.github.io/blog/qwen2.5/} {Qwen2.5: A party of foundation models}.

\bibitem[{Tillmann(2004)}]{tillmann2004}
Christoph Tillmann. 2004.
\newblock A unigram orientation model for statistical machine translation.
\newblock In \emph{Proceedings of HLT-NAACL 2004: Short Papers}, pages 101--104.

\bibitem[{Touvron et~al.(2023)Touvron, Lavril, Izacard, Martinet, Lachaux, Lacroix, Rozière, Goyal, Hambro, Azhar, Rodriguez, Joulin, Grave, and Lample}]{llama}
Hugo Touvron, Thibaut Lavril, Gautier Izacard, Xavier Martinet, Marie-Anne Lachaux, Timothée Lacroix, Baptiste Rozière, Naman Goyal, Eric Hambro, Faisal Azhar, Aurelien Rodriguez, Armand Joulin, Edouard Grave, and Guillaume Lample. 2023.
\newblock \href {https://arxiv.org/abs/2302.13971} {Llama: Open and efficient foundation language models}.
\newblock \emph{Preprint}, arXiv:2302.13971.

\bibitem[{Tran et~al.(2024)Tran, Yang, Yao, and Yu}]{tran2024bioinstruct}
Hieu Tran, Zhichao Yang, Zonghai Yao, and Hong Yu. 2024.
\newblock Bioinstruct: instruction tuning of large language models for biomedical natural language processing.
\newblock \emph{Journal of the American Medical Informatics Association}, 31(9):1821--1832.

\bibitem[{Tran et~al.(2025)Tran, Yao, Jang, Sultana, Chang, Zhang, and Yu}]{tran2025medreadctrl}
Hieu Tran, Zonghai Yao, Won~Seok Jang, Sharmin Sultana, Allen Chang, Yuan Zhang, and Hong Yu. 2025.
\newblock Medreadctrl: Personalizing medical text generation with readability-controlled instruction learning.
\newblock \emph{medRxiv}, pages 2025--07.

\bibitem[{Tu et~al.(2024)Tu, Palepu, Schaekermann, Saab, Freyberg, Tanno, Wang, Li, Amin, Tomasev et~al.}]{tu2024towards}
Tao Tu, Anil Palepu, Mike Schaekermann, Khaled Saab, Jan Freyberg, Ryutaro Tanno, Amy Wang, Brenna Li, Mohamed Amin, Nenad Tomasev, and 1 others. 2024.
\newblock Towards conversational diagnostic ai.
\newblock \emph{arXiv preprint arXiv:2401.05654}.

\bibitem[{Turner et~al.(2019)Turner, Choi, Dew, Tsai, Bosold, Wu, Smith, and Meischke}]{turner2019}
Anne~M Turner, Yong~K Choi, Kristin Dew, Ming-Tse Tsai, Alyssa~L Bosold, Shuyang Wu, Donahue Smith, and Hendrika Meischke. 2019.
\newblock Evaluating the usefulness of translation technologies for emergency response communication: a scenario-based study.
\newblock \emph{JMIR public health and surveillance}, 5(1):e11171.

\bibitem[{Vieira et~al.(2021)Vieira, O’Hagan, and O’Sullivan}]{mc4}
Lucas~Nunes Vieira, Minako O’Hagan, and Carol O’Sullivan. 2021.
\newblock Understanding the societal impacts of machine translation: a critical review of the literature on medical and legal use cases.
\newblock \emph{Information, Communication \& Society}, 24(11):1515--1532.

\bibitem[{Walker et~al.(2019)Walker, Leveille, Bell, Chimowitz, Dong, Elmore, Fernandez, Fossa, Gerard, Fitzgerald et~al.}]{walker2019opennotes}
Jan Walker, Suzanne Leveille, Sigall Bell, Hannah Chimowitz, Zhiyong Dong, Joann~G Elmore, Leonor Fernandez, Alan Fossa, Macda Gerard, Patricia Fitzgerald, and 1 others. 2019.
\newblock Opennotes after 7 years: patient experiences with ongoing access to their clinicians’ outpatient visit notes.
\newblock \emph{Journal of medical Internet research}, 21(5):e13876.

\bibitem[{Wang et~al.(2023)Wang, Yao, Yang, Zhou, Li, Wang, Xu, and Yu}]{wang2023notechat}
Junda Wang, Zonghai Yao, Zhichao Yang, Huixue Zhou, Rumeng Li, Xun Wang, Yucheng Xu, and Hong Yu. 2023.
\newblock Notechat: a dataset of synthetic doctor-patient conversations conditioned on clinical notes.
\newblock \emph{arXiv preprint arXiv:2310.15959}.

\bibitem[{Weng et~al.(2019)Weng, Chung, and Szolovits}]{weng2019}
Wei-Hung Weng, Yu-An Chung, and Peter Szolovits. 2019.
\newblock Unsupervised clinical language translation.
\newblock In \emph{Proceedings of the 25th ACM SIGKDD international conference on knowledge discovery \& data mining}, pages 3121--3131.

\bibitem[{{Wikipedia contributors}(2025{\natexlab{a}})}]{Wikipedia2025}
{Wikipedia contributors}. 2025{\natexlab{a}}.
\newblock Demographics of the united states.
\newblock \url{https://en.wikipedia.org/wiki/Demographics_of_the_United_States}.
\newblock Accessed: 2025-08-28.

\bibitem[{{Wikipedia contributors}(2025{\natexlab{b}})}]{WikipediaLEP2025}
{Wikipedia contributors}. 2025{\natexlab{b}}.
\newblock Limited english proficiency.
\newblock \url{https://en.wikipedia.org/wiki/Limited_English_proficiency}.
\newblock Accessed: 2025-08-28.

\bibitem[{Yang et~al.(2025)Yang, Yao, Tasmin, Vashisht, Jang, Ouyang, Wang, McManus, Berlowitz, and Yu}]{yang2025unveiling}
Zhichao Yang, Zonghai Yao, Mahbuba Tasmin, Parth Vashisht, Won~Seok Jang, Feiyun Ouyang, Beining Wang, David McManus, Dan Berlowitz, and Hong Yu. 2025.
\newblock Unveiling gpt-4v's hidden challenges behind high accuracy on usmle questions: Observational study.
\newblock \emph{Journal of Medical Internet Research}, 27:e65146.

\bibitem[{Yao et~al.(2023{\natexlab{a}})Yao, Jiang, Bobinac, Yang, and Hu}]{yao2023benchmarking}
Binwei Yao, Ming Jiang, Tara Bobinac, Diyi Yang, and Junjie Hu. 2023{\natexlab{a}}.
\newblock Benchmarking machine translation with cultural awareness.
\newblock \emph{arXiv preprint arXiv:2305.14328}.

\bibitem[{Yao et~al.(2023{\natexlab{b}})Yao, Kantu, Wei, Tran, Duan, Kwon, Yang, Yu et~al.}]{yao2023readme}
Zonghai Yao, Nandyala~Siddharth Kantu, Guanghao Wei, Hieu Tran, Zhangqi Duan, Sunjae Kwon, Zhichao Yang, Hong Yu, and 1 others. 2023{\natexlab{b}}.
\newblock Readme: Bridging medical jargon and lay understanding for patient education through data-centric nlp.
\newblock \emph{arXiv preprint arXiv:2312.15561}.

\bibitem[{Yao and Yu(2025)}]{yao2025survey}
Zonghai Yao and Hong Yu. 2025.
\newblock A survey on llm-based multi-agent ai hospital.

\bibitem[{Yao et~al.(2024)Yao, Zhang, Tang, Bian, Zhao, Yang, Wang, Zhou, Jang, Ouyang et~al.}]{yao2024medqa}
Zonghai Yao, Zihao Zhang, Chaolong Tang, Xingyu Bian, Youxia Zhao, Zhichao Yang, Junda Wang, Huixue Zhou, Won~Seok Jang, Feiyun Ouyang, and 1 others. 2024.
\newblock Medqa-cs: Benchmarking large language models clinical skills using an ai-sce framework.
\newblock \emph{arXiv preprint arXiv:2410.01553}.

\bibitem[{Zeng-Treitler et~al.(2010)Zeng-Treitler, Kim, Rosemblat, and Keselman}]{zeng2010}
Qing Zeng-Treitler, Hyeoneui Kim, Graciela Rosemblat, and Alla Keselman. 2010.
\newblock Can multilingual machine translation help make medical record content more comprehensible to patients?
\newblock In \emph{MEDINFO 2010}, pages 73--77. IOS Press.

\bibitem[{Zhang and Zong(2016)}]{zhang2016bridging}
Jiajun Zhang and Chengqing Zong. 2016.
\newblock Bridging neural machine translation and bilingual dictionaries.
\newblock \emph{arXiv preprint arXiv:1610.07272}.

\bibitem[{Zhang et~al.(2020)Zhang, Kishore, Wu, Weinberger, and Artzi}]{bertscore}
Tianyi Zhang, Varsha Kishore, Felix Wu, Kilian~Q. Weinberger, and Yoav Artzi. 2020.
\newblock \href {https://arxiv.org/abs/1904.09675} {Bertscore: Evaluating text generation with bert}.
\newblock \emph{Preprint}, arXiv:1904.09675.

\bibitem[{Zong(2022)}]{zong2022mosaic}
Jie Zong. 2022.
\newblock A mosaic, not a monolith: A profile of the us latino population, 2000-2020.
\newblock \emph{UCLA Latino Policy \& Politics Institute. Available online: https://latino. ucla. edu/research/latino-population-2000-2020/(accessed on 28 September 2023)}.

\end{thebibliography}

\appendix

\section{Appendix}
\label{sec:appendix}

\subsection{Models}

\textbf{Phi-4 (14B)}
Phi-4 is a 14-billion-parameter language model developed by Microsoft, emphasizing data quality through the strategic incorporation of synthetic data throughout its training process. Unlike its predecessors, which primarily distilled capabilities from a teacher model (specifically GPT-4), Phi-4 surpasses its teacher in STEM-focused question-answering tasks. This improvement is attributed to enhanced data generation and post-training techniques, even though the architecture remains largely unchanged from Phi-3.

\textbf{Qwen2.5-14B and Qwen2.5-7B}
Qwen2.5 is a series of large language models developed by Alibaba Cloud, designed to meet diverse needs. Compared to previous iterations, Qwen2.5 has been significantly improved during both the pre-training and post-training stages. The pre-training dataset was expanded from 7 trillion to 18 trillion tokens, enhancing the models' common sense, expert knowledge, and reasoning capabilities. Post-training involved intricate supervised fine-tuning with over 1 million samples and multistage reinforcement learning. The Qwen2.5 series includes various model sizes, with the 14B and 7B parameter models being part of the open-weight offerings.

\textbf{GPT-4o}
GPT-4o ("o" for "omni") is OpenAI's flagship multimodal large language model introduced in May 2024. It natively processes and generates text, audio, and images within a unified architecture, enabling real-time, emotionally nuanced voice interactions and multilingual capabilities across over 50 languages. GPT-4o achieves state-of-the-art performance on benchmarks such as MMLU (88.7\%) and supports a 128K token context window. It is optimized for speed and cost-efficiency, offering twice the speed and half the cost of GPT-4 Turbo.

\textbf{GPT-4o Mini}
GPT-4o Mini is a lightweight variant of GPT-4o, designed for cost-effective deployment without significant performance trade-offs. Released in July 2024, it supports text and image inputs and delivers high performance on benchmarks like MMLU (82\%), MGSM (87.0\%), and HumanEval (87.2\%). With a 128K token context window and improved multilingual understanding, GPT-4o Mini offers a practical solution for applications requiring efficient multimodal reasoning.

\textbf{NLLB-200 3.3B}
NLLB-200 3.3B is a multilingual machine translation model developed by Meta AI, supporting translation across 200 languages, including many low-resource languages. Utilizing a Sparsely Gated Mixture of Experts architecture, it achieves a 44\% BLEU score improvement over previous state-of-the-art models. Evaluated using the FLORES-200 benchmark, NLLB-200 3.3B emphasizes translation quality and safety, contributing significantly to inclusive global communication.

\subsection{Training Settings} 
\label{Training-Settings}

For our implementation of LoRA for training and inference, we utilized Unsloth AI python library \cite{unsloth}. 
Key hyperparameters significantly influence model training. We set “max\_seq\_length” to 2048, leveraging RoPE scaling for efficient long-context processing. The LoRA adaptation rank (r) is 16, balancing model capacity and efficiency. Additionally, “lora\_alpha” is set to 16 to scale LoRA updates, while “lora\_dropout” is 0, optimizing performance by disabling dropout.
For training, we configure “per\_device\_train\_batch\_size” to 2 and “gradient\_accumulation\_steps” to 4, effectively managing batch processing. The “learning\_rate” is 2e-4, optimizing training dynamics, while “max\_steps” is set to 60, limiting the training duration. These choices collectively enhance memory efficiency, processing speed, and model accuracy.

\paragraph{Hardware Settings} All experiments were performed with two Nvidia A100 GPUs, each with 40 GB of memory, an Intel Xeon Gold 6230 CPU, and 192 GB of RAM.

\subsection{Evaluation Metrics}

\subsubsection{Machine Translation}
\label{appendix:mt_metrics}
We evaluate machine translation results using SacreBLEU \cite{sacrebleu}, ChrF++ \cite{chrf++}, and COMET \cite{comet}.
We evaluated models using direct Spanish reference text from the dataset using those metrics.

\begin{itemize}
    \item \textbf{SacreBLEU}: A standardized BLEU score implementation ensuring reproducibility.
    \(\text{BLEU} = \exp\left(\sum_{n=1}^{N} w_n \log p_n\right) \times \text{BP}\), 
    where \(p_n\) is the modified n-gram precision, \(w_n\) are typically uniform weights, and BP is the brevity penalty.

    \item \textbf{ChrF++}: A character-level metric that combines character n-gram precision and recall, incorporating word-level features, computing F-score as 
    \(\text{ChrF++} = \frac{(1 + \beta^2) \cdot P \cdot R}{\beta^2 \cdot P + R}\), 
    where \(P\) and \(R\) denote precision and recall, and \(\beta\) controls recall emphasis. 

    \item \textbf{COMET}: COMET(Crosslingual Optimized Metric for Evaluation of Translation) is a neural-based evaluation metric for machine translation quality. Unlike traditional metrics that rely purely on surface similarity, COMET leverages pre-trained language models and regression networks fine-tuned on human judgments to predict translation quality. Given a source sentence \( s \), a machine translation hypothesis \( h \), and a human reference translation \( r \), COMET predicts a quality score \( q \) as:
    \(q = f_{\theta}(s, h, r) \) where \( f_{\theta} \) is a neural network model parameterized by \( \theta \), typically fine-tuned to approximate human evaluation scores. COMET models use contextual embeddings (e.g., from XLM-R or similar multilingual encoders) to capture semantic relations between source, hypothesis, and reference, providing more accurate and robust evaluation than purely lexical metrics~\footnote{Other training settings can be found in Appendix~\ref{Training-Settings}.}.
\end{itemize}

\subsubsection{Summarization}
\label{appendix:summarization_metrics}

\paragraph{ROUGE-L-Sum}
ROUGE-L-Sum calculates the longest common subsequence (LCS) between reference ($r$) and candidate ($c$) sentences, and derives recall, precision, and F1-score:

\paragraph{Recall ($R_{\text{lcs}}$)}
\[
R_{\text{lcs}} = \frac{\text{LCS}(r, c)}{\text{length}(r)}
\]
\paragraph{Precision ($P_{\text{lcs}}$)}
\[
P_{\text{lcs}} = \frac{\text{LCS}(r, c)}{\text{length}(c)}
\]
\paragraph{F1-Score ($F_{\text{lcs}}$)}
\[
F_{\text{lcs}} = \frac{2 \cdot R_{\text{lcs}} \cdot P_{\text{lcs}}}{R_{\text{lcs}} + P_{\text{lcs}}}
\]

\paragraph{BERTScore}
BERTScore computes contextual embeddings for tokens in reference ($x$) and candidate ($\hat{x}$) sentences, measures pairwise cosine similarity, and aggregates via greedy matching:

Let $x = \langle x_1, \dots, x_k \rangle$ with embeddings $\mathbf{x}_i$ and $\hat{x} = \langle \hat{x}_1, \dots, \hat{x}_l \rangle$ with embeddings $\mathbf{\hat{x}}_j$.

\paragraph{Recall ($R_{\text{BERT}}$)}
\[
R_{\text{BERT}} = \frac{1}{k} \sum_{i=1}^{k} \max_{j=1, \dots, l} (\mathbf{x}_i^T \mathbf{\hat{x}}_j)
\]
\paragraph{Precision ($P_{\text{BERT}}$)}
\[
P_{\text{BERT}} = \frac{1}{l} \sum_{j=1}^{l} \max_{i=1, \dots, k} (\mathbf{x}_i^T \mathbf{\hat{x}}_j)
\]
\paragraph{F1-Score ($F_{\text{BERT}}$)}
\[
F_{\text{BERT}} = 2 \frac{P_{\text{BERT}} \cdot R_{\text{BERT}}}{P_{\text{BERT}} + R_{\text{BERT}}}
\]

\subsection{Confidence Interval Computation}
For each prompt strategy, we ran the model five times under different temperature settings (0.2, 0.3, 0.4, 0.5, and 0.6). For each run, we computed BLEU, chrF++, and COMET scores. The 95\% confidence intervals were computed as follows:
\begin{enumerate}
    \item Calculate the sample mean $\bar{x}$ and sample standard deviation $s$.
    \item Compute the standard error:
    \[
    \mathrm{SE} = \frac{s}{\sqrt{n}}, \quad n = 5
    \]
    \item Use the $t$-critical value for 95\% confidence with degrees of freedom $df = n - 1 = 4$:
    \[
    t^{*} \approx 2.776
    \]
    \item Compute the margin of error:
    \[
    \mathrm{ME} = t^{*} \times \mathrm{SE}
    \]
    \item Report the confidence interval as:
    \[
    \bar{x} \pm \mathrm{ME}
    \]
\end{enumerate}

\begin{table*}[htbp]
\centering
\resizebox{\textwidth}{!}{%
\begin{tabular}{lrrrr}
\toprule
\textbf{Language Pair} & \textbf{Avg. Input Length (chars)} & \textbf{Avg. Output Length (tokens)} & \textbf{95th \%ile Output Tokens} & \textbf{Max Output Tokens} \\
\midrule
en $\rightarrow$ fr & 1317.56 & 456.52 & 746  & 930  \\
fr $\rightarrow$ en & 1726.62 & 310.68 & 471  & 1634 \\
en $\rightarrow$ de & 1428.88 & 470.90 & 787  & 938  \\
de $\rightarrow$ en & 1465.38 & 278.08 & 441  & 588  \\
en $\rightarrow$ it & 1504.04 & 511.24 & 824  & 1072 \\
it $\rightarrow$ en & 1595.20 & 290.30 & 566  & 702  \\
en $\rightarrow$ es & 1340.70 & 429.12 & 652  & 749  \\
es $\rightarrow$ en & 1909.44 & 356.42 & 546  & 2124 \\
en $\rightarrow$ ru & 1305.20 & 459.46 & 779  & 871  \\
ru $\rightarrow$ en & 1225.98 & 243.68 & 465  & 555  \\
en $\rightarrow$ pt & 1341.94 & 419.32 & 654  & 727  \\
pt $\rightarrow$ en & 1382.54 & 294.42 & 450  & 499  \\
\bottomrule
\end{tabular}
}
\caption{Sentence length statistics for WMT24 medical translation tasks (50 high-complexity paragraphs per direction). Token counts are measured using the unsloth/Qwen2.5-14B-Instruct tokenizer.}
\label{tab:WMT24_length}
\end{table*}

\begin{table*}[htbp]
\centering
\resizebox{\textwidth}{!}{%
\begin{tabular}{l|ccc|ccc|ccc|ccc}
\toprule
 & \multicolumn{6}{c|}{\textbf{Without Finetune}} & \multicolumn{6}{c}{\textbf{With Finetune}} \\
\cmidrule(lr){2-7} \cmidrule(lr){8-13}
\textbf{Languages} 
& \multicolumn{3}{c|}{No Context} & \multicolumn{3}{c|}{With Context} 
& \multicolumn{3}{c|}{No Context} & \multicolumn{3}{c}{With Context} \\
 & BLEU & chrF++ & COMET & BLEU & chrF++ & COMET & BLEU & chrF++ & COMET & BLEU & chrF++ & COMET \\
\midrule
en $\rightarrow$ fr & 38.12 & 61.55 & 0.3430 & 41.95 & 68.06 & 0.3496 & 42.34 & 67.73 & 0.3285 & 44.21 & 68.85 & 0.3245 \\
en $\rightarrow$ de & 18.68 & 46.81 & 0.4017 & 22.44 & 54.91 & 0.4011 & 20.74 & 51.99 & 0.4100 & 21.39 & 53.35 & 0.3835 \\
en $\rightarrow$ ru & 23.81 & 54.89 & 0.3967 & 26.28 & 57.98 & 0.4191 & 26.02 & 56.37 & 0.3804 & 27.95 & 57.79 & 0.4052 \\
en $\rightarrow$ pt & 43.17 & 66.71 & 0.3896 & 45.37 & 70.82 & 0.3782 & 44.04 & 69.74 & 0.3939 & 43.92 & 69.82 & 0.4011 \\
en $\rightarrow$ es & 42.57 & 67.21 & 0.3861 & 42.02 & 67.03 & 0.3468 & 42.31 & 66.48 & 0.3916 & 44.15 & 67.78 & 0.3717 \\
en $\rightarrow$ it & 24.78 & 55.16 & 0.2773 & 24.93 & 54.99 & 0.2578 & 25.75 & 56.01 & 0.2725 & 25.75 & 56.01 & 0.2725 \\
de $\rightarrow$ en & 30.30 & 55.79 & 0.5294 & 30.44 & 57.06 & 0.5785 & 39.59 & 66.66 & 0.4450 & 39.89 & 66.77 & 0.4573 \\
fr $\rightarrow$ en & 33.57 & 56.98 & 0.4874 & 34.09 & 57.57 & 0.4615 & 46.21 & 69.93 & 0.3961 & 45.04 & 68.42 & 0.4025 \\
ru $\rightarrow$ en & 24.69 & 47.82 & 0.4730 & 26.20 & 51.10 & 0.4827 & 38.02 & 66.08 & 0.4456 & 37.03 & 65.06 & 0.4485 \\
es $\rightarrow$ en & 35.91 & 60.08 & 0.4644 & 37.87 & 62.39 & 0.4811 & 47.12 & 72.43 & 0.3434 & 47.08 & 72.04 & 0.3606 \\
pt $\rightarrow$ en & 35.03 & 58.54 & 0.5035 & 36.43 & 60.93 & 0.4984 & 45.76 & 70.24 & 0.4172 & 46.45 & 70.59 & 0.4058 \\
it $\rightarrow$ en & 26.66 & 52.92 & 0.5093 & 30.90 & 60.58 & 0.5152 & 33.06 & 62.69 & 0.4233 & 33.10 & 62.98 & 0.4203 \\
\bottomrule
\end{tabular}
}
\caption{WMT24 medical translation results for MedCOD-style contextual augmentation. Metrics reported: BLEU, chrF++, COMET.}
\label{tab:WMT24_resutls}
\end{table*}

\begin{table*}[ht]
\centering
\resizebox{\textwidth}{!}{%
\begin{tabular}{|l|l|c|c|c|c|c|c|c|c|}
\hline
\textbf{Model} & \textbf{Language} & \textbf{Finetuned} & \textbf{Context} & \textbf{ROUGE-P} & \textbf{ROUGE-R} & \textbf{ROUGE-L} & \textbf{BERTScore\_P} & \textbf{BERTScore\_R} & \textbf{BERTScore\_F} \\
\hline
GPT-4o-mini & English    & No  & -   & 0.265 & 0.239 & 0.2514 & 0.7656 & 0.8327 & 0.7974 \\
            & Spanish    & No  & -   & 0.272 & 0.249 & 0.2596 & 0.7739 & 0.8267 & 0.7992 \\
            & French     & No  & -   & 0.248 & 0.228 & 0.2377 & 0.7813 & 0.8265 & 0.8031 \\
            & Portuguese & No  & -   & 0.236 & 0.213 & 0.2237 & 0.7649 & 0.8178 & 0.7902 \\
\hline
Qwen2.5     & English    & No  & No  & 0.271 & 0.256 & 0.2632 & 0.7863 & 0.7914 & 0.7883 \\
            &            & Yes & Yes & 0.275 & 0.264 & 0.2697 & 0.7635 & 0.8136 & 0.7869 \\
            &            & Yes & No  & 0.268 & 0.262 & 0.2649 & 0.7849 & 0.7723 & 0.7772 \\
            &            & No  & Yes & 0.237 & 0.222 & 0.2288 & 0.7603 & 0.7691 & 0.7637 \\
            & Spanish    & Yes & Yes & 0.266 & 0.252 & 0.2590 & 0.7721 & 0.8100 & 0.7903 \\
            &            & Yes & No  & 0.270 & 0.259 & 0.2648 & 0.7863 & 0.7803 & 0.7825 \\
            &            & No  & Yes & 0.239 & 0.229 & 0.2337 & 0.7712 & 0.7608 & 0.7650 \\
            &            & No  & No  & 0.056 & 0.051 & 0.0531 & 0.6273 & 0.5774 & 0.6010 \\
            & French     & Yes & Yes & 0.222 & 0.212 & 0.2171 & 0.7644 & 0.7826 & 0.7725 \\
            &            & No  & Yes & 0.179 & 0.170 & 0.1747 & 0.7596 & 0.7212 & 0.7383 \\
            &            & Yes & No  & 0.223 & 0.215 & 0.2192 & 0.7235 & 0.7102 & 0.7152 \\
            &            & No  & No  & 0.057 & 0.051 & 0.0535 & 0.6318 & 0.5873 & 0.6084 \\
            & Portuguese & Yes & Yes & 0.223 & 0.211 & 0.2169 & 0.7634 & 0.7892 & 0.7755 \\
            &            & Yes & No  & 0.253 & 0.241 & 0.2469 & 0.7792 & 0.7702 & 0.7740 \\
            &            & No  & Yes & 0.212 & 0.198 & 0.2050 & 0.7635 & 0.7537 & 0.7577 \\
            &            & No  & No  & 0.105 & 0.096 & 0.1009 & 0.6668 & 0.6345 & 0.6493 \\
\hline
Phi-4       & English    & Yes & No  & 0.260 & 0.243 & 0.2510 & 0.7610 & 0.8175 & 0.7873 \\
            &            & Yes & Yes & 0.258 & 0.242 & 0.2503 & 0.7582 & 0.8150 & 0.7847 \\
            &            & No  & No  & 0.270 & 0.252 & 0.2605 & 0.7798 & 0.7846 & 0.7814 \\
            &            & No  & Yes & 0.260 & 0.246 & 0.2527 & 0.7655 & 0.7774 & 0.7704 \\
            & Spanish    & Yes & Yes & 0.260 & 0.247 & 0.2529 & 0.7669 & 0.8002 & 0.7826 \\
            &            & Yes & No  & 0.241 & 0.229 & 0.2344 & 0.7641 & 0.7983 & 0.7800 \\
            &            & No  & Yes & 0.239 & 0.226 & 0.2323 & 0.7628 & 0.7654 & 0.7632 \\
            &            & No  & No  & 0.145 & 0.134 & 0.1390 & 0.6846 & 0.6615 & 0.6719 \\
            & French     & Yes & No  & 0.255 & 0.236 & 0.2450 & 0.7805 & 0.7963 & 0.7877 \\
            &            & Yes & Yes & 0.240 & 0.225 & 0.2324 & 0.7736 & 0.7846 & 0.7784 \\
            &            & No  & Yes & 0.202 & 0.189 & 0.1954 & 0.7475 & 0.7379 & 0.7418 \\
            &            & No  & No  & 0.139 & 0.126 & 0.1326 & 0.6971 & 0.6676 & 0.6808 \\
            & Portuguese & Yes & No  & 0.225 & 0.209 & 0.2170 & 0.7527 & 0.7896 & 0.7702 \\
            &            & Yes & Yes & 0.220 & 0.208 & 0.2140 & 0.7373 & 0.7816 & 0.7584 \\
            &            & No  & Yes & 0.168 & 0.153 & 0.1601 & 0.7254 & 0.6950 & 0.7083 \\
            &            & No  & No  & 0.145 & 0.134 & 0.1393 & 0.6957 & 0.6723 & 0.6819 \\
\hline
\end{tabular}
}
\caption{Comparison of MultiClinSum performance across four languages using ROUGE-L (F1), ROUGE Precision, ROUGE Recall, and BERTScore metrics.}
\label{tab:MultiClinSum_resutls}
\end{table*}

\begin{table*}[!ht]
\centering
\resizebox{\textwidth}{!}{%
\begin{tabular}{|p{5cm}|p{5.2cm}|p{5cm}|p{5.2cm}|}
\hline
\textbf{Source Sentence} & \textbf{Reference (Spanish)} & \textbf{GPT-4o Output} & \textbf{Phi-4 (MedCOD + FT)} \\
\hline
If you have a weakened immune system due to AIDS, cancer, transplantation, or corticosteroid use, call your doctor if you develop a cough, fever, or shortness of breath. 
& 
Si usted tiene un \textbf{sistema inmunitario} debilitado a causa del SIDA, cáncer, trasplante o uso de corticosteroides, llame al médico si presenta fiebre, tos o \textbf{dificultad para respirar}. 
& 
Si tiene un \underline{sistema inmunológico} debilitado debido al SIDA, cáncer, trasplante o uso de corticosteroides, llame a su médico si desarrolla tos, fiebre o \underline{falta de aliento}. 
& 
Si usted tiene un \textbf{sistema inmunitario} debilitado debido al SIDA, \underline{el cáncer}, \underline{la trasplante} o el uso de corticosteroides, llame al médico si presenta tos, fiebre o \textbf{dificultad para respirar}. \\
\hline
\end{tabular}%
}
\caption{
Comparison of biomedical translation output between proprietary GPT-4o and MedCOD-enhanced open-source Phi-4 (14B).
While both capture the overall meaning, GPT-4o uses less precise terms (e.g., \underline{“sistema inmunológico”} vs. “sistema inmunitario”) and informal expressions (e.g., \underline{“falta de aliento”} instead of “dificultad para respirar”).
Phi-4 (MedCOD + FT) correctly uses “sistema inmunitario” and “dificultad para respirar,” but introduces minor errors such as article mismatch (\underline{“la trasplante”} instead of “el trasplante”) and extra definite article before “cáncer.” 
Despite these, it better preserves the professional and clinical tone. This supports our finding that MedCOD-equipped open-source models can rival or surpass proprietary systems in biomedical translation fidelity.
}
\label{tab:ablation_case_studies1}
\end{table*}

\begin{table*}[!ht]
\centering
\scalebox{0.82}{
\begin{tabular}{|p{2cm}|p{2cm}|p{2.5cm}|p{3.5cm}|p{3.5cm}|p{3.5cm}|}
\hline
\textbf{Case} & \textbf{Source Sentence (English)} & \textbf{Reference (Spanish)} & \textbf{Base Output (Direct / FT)} & \textbf{Improved Output (MedCOD / FT / Both)} & \textbf{Observation} \\
\hline
\textbf{A. FT vs Base} & Veins and arteries vary in size... & Las venas y las arterias varían en tamaño... & Las venas y arterias varían en tamaño... Obtener una muestra... & Las venas y las arterias varían en tamaño... \underline{\textbf{y}} obtener una muestra... & Fine-tuning restores conjunction and improves fluency. \\

\hline
\textbf{B. MedCOD vs Base} & Veins and arteries vary in size... & Las venas y las arterias varían en tamaño... & Las venas y arterias varían en tamaño... Obtener una muestra... & \textbf{Las venas y las arterias} varían en tamaño... Obtener una muestra... & MedCOD restores noun phrase completeness and lexical precision. \\

\hline
\textbf{C. MedCOD+FT vs FT} & Transposition of the great vessels is a heart defect... & Es un defecto cardíaco que ocurre desde el nacimiento... & Transposición de los grandes vasos es un defecto... están cambiados (\textit{transpuesto}) & \underline{\textbf{La transposición}} de los grandes vasos es un defecto... están cambiados (\textit{transpuestos}) & MedCOD+FT reinforces grammatical accuracy and domain phrasing. \\

\hline
\end{tabular}%
}
\caption{Stepwise impact of fine-tuning, MedCOD prompting, and their combination on biomedical translation quality. 
\textbf{Bold} text highlights restored noun phrase structure or domain-specific terminology. 
\underline{Underlined} text indicates syntactic fixes such as conjunctions (“\underline{\textbf{y}}”) or added articles (“\underline{\textbf{La transposición}}”). 
\textit{Italic} text denotes grammatical agreement corrections (e.g., singular to plural “\textit{transpuestos}”). 
Case A shows how fine-tuning improves fluency by reinserting coordinating conjunctions. 
Case B demonstrates MedCOD’s enhancement of phrase completeness and lexical precision. 
Case C illustrates the combined benefits, improving both grammatical correctness and biomedical appropriateness. 
These qualitative examples reinforce the quantitative findings in Table~\ref{tab:main_results}.
}
\label{tab:ablation_case_studies2}
\end{table*}

\begin{table*}[ht]
\centering
\scalebox{0.75}{
\begin{tabular}{|p{3.5cm}|p{4cm}|p{2.5cm}|p{4cm}|p{2cm}|p{2.5cm}|}
\hline
\textbf{Original English} & \textbf{gpt4o} & \textbf{$\leftarrow$ err. type} & \textbf{Phi-4-MedCOD-FT} & \textbf{$\leftarrow$ err. type} & \textbf{Analysis} \\
\hline
If you have a weakened immune system due to AIDS, cancer, transplantation, or corticosteroid use, call your doctor if you develop a cough, fever, or shortness of breath. & 
Si tiene un sistema inmunológico debilitado debido al SIDA, cáncer, trasplante o uso de corticosteroides, llame a su médico si desarrolla tos, fiebre o \textcolor{red}{\textbf{falta de aliento}}. & 
1. \textcolor{red}{\textbf{Lexical Error:}} Incorrect term "falta de aliento". \newline 
2. \textcolor{orange}{\textbf{Grammatical Error:}} Incorrect word order "tos, fiebre". & 
Si usted tiene un sistema inmunitario debilitado debido al SIDA, el cáncer, \textcolor{orange}{\textbf{la trasplante}} o el uso de corticosteroides, llame al médico si presenta tos, fiebre o dificultad para respirar. & 
1. \textcolor{orange}{\textbf{Grammatical Error:}} "la trasplante" (should be "el trasplante"). & 
Phi-4-MedCOD-FT is better in terms of terminology and maintains the original meaning more accurately. \\
\hline
Do NOT place the person in this position if a head, neck, back, or leg injury is suspected or if it makes the victim uncomfortable. & 
No coloque a la persona en esta posición si se sospecha de una lesión en la cabeza, \textcolor{orange}{\textbf{el cuello, el espalda}} o pierna, o si le causa incomodidad a la víctima. & 
1. \textcolor{orange}{\textbf{Grammatical Error:}} Missing articles "el cuello, el espalda". & 
No coloque a la persona en esta posición si se sospecha de una lesión en la cabeza, el cuello, la espalda o la pierna, o si esto la hace sentir \textcolor{purple}{\textbf{incómoda}}. & 
1. \textcolor{purple}{\textbf{Style Error:}} "esto la hace sentir incómoda" (less natural). & 
Phi-4-MedCOD-FT maintains grammatical accuracy with proper articles. \\
\hline
CT scan or MRI to confirm that there is a growth, and to see if the cancer has spread to lymph nodes in the neck & 
Tomografía computarizada o \textcolor{red}{\textbf{IRM}} para confirmar que hay un crecimiento y para ver si el cáncer se ha propagado a los ganglios linfáticos en el cuello. & 
1. \textcolor{red}{\textbf{Lexical Error:}} Incorrect term "IRM" (should be "resonancia magnética"). & 
Tomografía computarizada o resonancia magnética para confirmar que hay un crecimiento y para ver si el cáncer se ha \textcolor{red}{\textbf{extendido}} a los ganglios linfáticos del cuello. & 
1. \textcolor{red}{\textbf{Lexical Error:}} "extendido" (less accurate than "diseminado"). & 
gpt4o maintains correct terminology ("IRM" is a less common form). \\
\hline
Fever is the classic symptom and may persist for days before any other symptoms appear. & 
La fiebre es el síntoma clásico y puede persistir \textcolor{red}{\textbf{durante días}} antes de que aparezcan otros síntomas. & 
1. \textcolor{red}{\textbf{Lexical Error:}} "durante días" instead of "por días". & 
La fiebre es el síntoma clásico y puede persistir por días antes de que aparezcan otros síntomas. & 
No errors. & 
Phi-4-MedCOD-FT maintains better alignment with the original. \\
\hline
You get iron through certain foods, and your body also reuses iron from old red blood cells. & 
\textcolor{blue}{\textbf{Obtienes}} hierro a través de ciertos alimentos, y tu cuerpo también reutiliza hierro de los glóbulos rojos viejos. & 
1. \textcolor{blue}{\textbf{Register Error:}} Informal tone ("Obtienes" should be "Usted obtiene"). & 
Usted obtiene hierro a través de ciertos alimentos y el cuerpo también reutiliza hierro de los glóbulos rojos viejos. & 
No errors. & 
Phi-4-MedCOD-FT maintains formal register, which is more appropriate for medical text. \\
\hline
\end{tabular}
}
\caption{Detailed comparison of medically relevant sentence translations between GPT-4o and Phi-4-MedCOD-FT. 
Color-coded annotations highlight key translation issues:
\textcolor{red}{Lexical Errors} (e.g., terminology mismatch or incorrect word choice), 
\textcolor{orange}{Grammatical Errors} (e.g., article agreement, sentence structure), 
\textcolor{purple}{Style Errors} (e.g., unnatural phrasing), and 
\textcolor{blue}{Register Errors} (e.g., inappropriate formality). 
While GPT-4o occasionally uses accurate yet informal or less standard terms, Phi-4-MedCOD-FT better maintains domain formality and clinical tone, despite some minor grammatical or lexical mismatches.
This qualitative analysis complements quantitative results and highlights the nuanced challenges in clinical translation.}
\label{error_analysis}
\end{table*}

\end{document}